%% file: main.tex
\documentclass[10pt,twocolumn,letterpaper]{article}

\usepackage[pagenumbers]{wacv} 

\input{preamble}
\input{macro}

\definecolor{wacvblue}{rgb}{0.21,0.49,0.74}
\usepackage[pagebackref,breaklinks,colorlinks,allcolors=wacvblue]{hyperref}

\def\wacvPaperID{1284} 
\def\confName{WACV} 
\def\confYear{2027}

\title{Relation-Centric Open-Vocabulary 3D Gaussian Segmentation}

\author{
  Eunsung Cha\footnotemark[1]\quad\quad
  Hyunjoon Lee\footnotemark[1]\quad\quad
  Jaesik Park\footnotemark[2]\\
  Seoul National University, Republic of Korea\\
  {\tt\small \{ender3553, hjlee4772, jaesik.park\}@snu.ac.kr} 
}

\begin{document}
\maketitle

\renewcommand{\thefootnote}{\fnsymbol{footnote}}
\footnotetext[1]{Equal Contribution.}
\footnotetext[2]{Corresponding author.}

\input{figs/teaser}
\input{sec/0_abstract}
\input{sec/1_intro}
\input{sec/2_related}
\input{sec/3_method}

\input{sec/4_experiments}
\input{sec/5_conclusion}

{
    \small
    \bibliographystyle{ieeenat_fullname}
    \bibliography{main}
}
\input{sec/X_suppl}
\end{document}

%% file: preamble.tex
\usepackage{multirow}
\usepackage{algorithm}
\usepackage[noend]{algpseudocode}
\algrenewcommand{\algorithmicindent}{1.0em}
\usepackage[hang,flushmargin,symbol]{footmisc}
\usepackage{graphicx}	
\usepackage{amsmath}	
\usepackage{amssymb}	
\usepackage{booktabs}
\usepackage{microtype}
\usepackage{epsfig}
\usepackage{caption}
\usepackage{float}
\usepackage{placeins}
\usepackage{color, colortbl}
\usepackage{stfloats}
\usepackage{enumitem}
\usepackage{tabularx}
\usepackage{xstring}
\usepackage{xspace}
\usepackage{url}
\usepackage{subcaption}
\usepackage{xcolor}
\usepackage{pifont}
\usepackage{cuted}  
\usepackage{caption}
\usepackage[table]{xcolor} 
\usepackage{array}
\usepackage{wrapfig}
\usepackage{array}
\newcolumntype{C}[1]{>{\centering\arraybackslash}p{#1}}
\usepackage{siunitx}
\usepackage{etoc}

%% file: macro.tex
\newcommand{\Ours}{PairGS}
\newcommand{\tree}{TreeDBSCAN}

\definecolor{colorf}{HTML}{FFC1C3}
\definecolor{colors}{HTML}{FFC991}
\definecolor{colort}{HTML}{FFF6A9}
\definecolor{softred}{HTML}{d62728}
\definecolor{softgray}{HTML}{A9A9A9}

\definecolor{softblue}{HTML}{1f77b4}
\definecolor{softgreen}{HTML}{2ca02c}
\definecolor{softpurple}{HTML}{9467bd}

\definecolor{colorf}{HTML}{FFC1C3}
\definecolor{colors}{HTML}{FFC991}
\definecolor{colort}{HTML}{FFF6A9}

%% file: figs/teaser.tex

\begin{strip}
\vspace*{-11mm}
\begin{center}
    \includegraphics[width=0.95\textwidth]{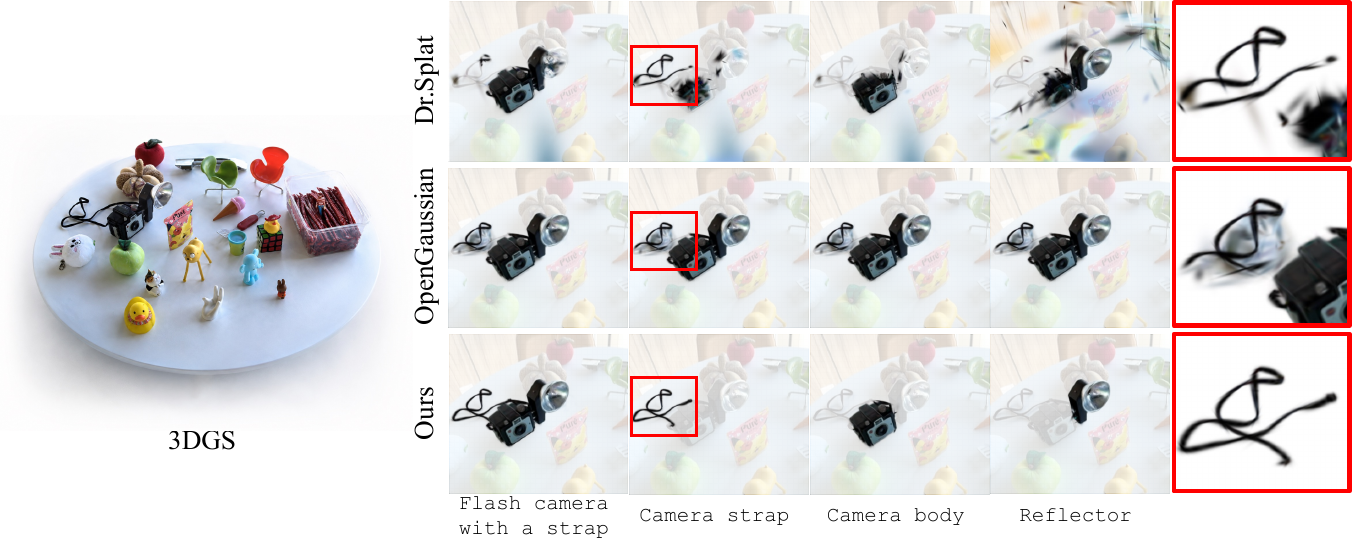}
    \captionof{figure}{We propose {\Ours}, a training-free method that estimates pairwise relations between Gaussians to enable efficient and fine-grained instance separation. {\Ours} supports multi-granular queries and produces cleaner separation even for thin parts that touch the floor, such as \texttt{camera strap}, compared with the language-based method (Dr.Splat~\cite{drsplat_2025}) and the instance-feature-based method (OpenGaussian~\cite{opengaussian_2024}).}
    \label{fig:teaser}
\end{center}
\vspace*{4mm}
\end{strip}

%% file: sec/0_abstract.tex
\begin{abstract}
Open-vocabulary 3D Gaussian segmentation is challenging because it requires language understanding for diverse queries and accurate separation of Gaussians along object boundaries.
Prior approaches either embed language knowledge into individual Gaussians to improve query responsiveness or optimize per-Gaussian instance features to encode object identity.
However, these strategies may produce noisy Gaussian segmentations or rely on cost-inefficient per-scene optimization.
We propose {\Ours}, a novel framework that reframes Gaussian segmentation as modeling pairwise relations between Gaussians. 
Our key insight is that 3D Gaussian representation inherently provides rich signals for relation estimation, such as view contribution weights and multi-view mask evidence. By leveraging these cues, we can explicitly construct a relation graph for segmentation without a heavy optimization process.
To construct the relation graph efficiently, we adopt a two-stage strategy that proposes sparse edge candidates using low-dimensional descriptors and then computes precise pairwise affinities only on these candidates.
From the resulting relation graph, {\Ours} constructs a hierarchical cluster tree.
This design supports various multi-granular queries while producing clean instance separation.
{\Ours} achieves state-of-the-art results on open-vocabulary 3D Gaussian segmentation benchmarks. A fast variant of {\Ours} is $50\times$ faster than optimization-based instance-feature approaches.

\end{abstract}

%% file: sec/1_intro.tex
\section{Introduction}
\label{sec:intro}
Recently, 3D Gaussian Splatting (3DGS)~\cite{3dgs_2023} has rapidly gained popularity due to its remarkable ability to achieve high-quality reconstruction and real-time rendering simultaneously. Its explicit representation and fast optimization scheme have made it a powerful foundation for various 3D tasks. Capitalizing on these strengths, a significant research trend has been to lift rich semantic features from 2D visual foundation models~\cite{clip_2021,sam_2023,dino_2021} into the 3DGS representation~\cite{langsplat_2024, drsplat_2025, feature3dgs_2024,ludvig_2024,cf3_2025,goi_2024,fmgs_2025,vlgs_2024, legaussian_2024}. Despite this progress, open-vocabulary segmentation in 3DGS scenes remains challenging.
It must support diverse language queries while separating fine-grained object instances with clean boundaries.

Two main paradigms have emerged for open-vocabulary segmentation in 3D Gaussian scenes: (1) language-based methods and (2) instance-feature-based methods.
Language-based methods~\cite{fmgs_2025, langsplat_2024, feature3dgs_2024,goi_2024,vlgs_2024,legaussian_2024, drsplat_2025, ludvig_2024, occams_2024, cf3_2025} distill 2D semantic features into individual Gaussians.
They segment the scene by computing query relevance scores for each Gaussian.
This design naturally supports diverse open-vocabulary queries.
However, since the scores are computed independently for each Gaussian, it does not explicitly enforce instance-level consistency across Gaussians.
As shown in \cref{fig:teaser}, the predicted instances are incoherent.

Instance-feature-based methods~\cite{opengaussian_2024, instancegaussian_2025, laga_2025, cos3d_2025, clickguassian_2024, saga_2025, gaussiangrouping_2024,gaga_2024,objectgs_2025} take a different route.
They optimize per-Gaussian instance features using 2D multi-view masks.
These features encourage Gaussians within the same object to share a common identity, thereby improving instance coherence.
However, the approach introduces two challenges.
First, it requires per-scene optimization, which can be computationally expensive.
Second, precise separation remains difficult in contact regions or for thin structures.
A key reason is that supervision is typically applied through a rendered instance feature map.
As discussed in prior works~\cite{occams_2024, ludvig_2024, drsplat_2025,cf3_2025}, this signal can be indirect for resolving 3D instance boundaries.
In addition, existing methods~\cite{opengaussian_2024, instancegaussian_2025, laga_2025, clickguassian_2024, saga_2025, gaussiangrouping_2024,gaga_2024,objectgs_2025} first produce a class-agnostic partition of the scene and later attach language semantics.
This can limit how the segmentation adapts to multi-granular queries, as illustrated in \cref{fig:teaser}.

To address these limitations, we introduce {\Ours}.
{\Ours} does not rely on \emph{per-Gaussian} language scores or \emph{per-scene} instance feature optimization.
Instead, we focus on \emph{pairwise} relations between Gaussians for the segmentation task.
Our key insight is that 3D Gaussian representation intrinsically offers sufficient cues for relation estimation.
It exposes view-dependent contribution weights and visibility across views.
Together with multi-view mask evidence, these cues allow for the explicit estimation of same-instance affinity between Gaussians.
{\Ours} builds a pairwise relation graph where each Gaussian is a node.
This design is training-free and avoids per-scene optimization.
It also enables fine-grained separation near object boundaries, as shown in~\cref{fig:teaser}.

\input{figs/overview}

{\Ours} constructs the relation graph in two stages, as shown in \cref{fig:overview}, which makes graph construction computationally efficient.
In the first stage, we build low-dimensional node descriptors and use them only to propose sparse edge candidates.
In the second stage, we compute pairwise affinities only on these candidates by aggregating same-instance evidence across views.
This design avoids exhaustive pairwise computation over all Gaussians.
Finally, we propose TreeDBSCAN, which organizes the relation graph into a single cohesive hierarchical cluster tree with explicit parent-child relations.
The tree yields a hierarchical clustering, enabling 3D Gaussian segmentation for multi-granular queries within a single unified structure.


We summarize our main contributions as follows:
\begin{itemize}
    \item \textbf{Relation-centric approach.} We introduce a 3D Gaussian segmentation framework that estimates pairwise relations between Gaussians, enabling clean separation even for thin structures and contact regions.
    \item \textbf{Efficient two-stage graph construction.} 
    We propose sparse edge candidates with low-dimensional descriptors, then compute expensive pairwise affinities only on these candidates.
    \item \textbf{A single hierarchy}: A weighted sparse digraph structure yields a cohesive hierarchical cluster tree, supporting multi-granular and fine-grained segmentation.
    \item \textbf{State-of-the-Art results}: {\Ours} achieves outstanding performance while being over $50\times$ faster than optimization-based methods with our fast variant.
\end{itemize}

%% file: figs/overview.tex
\begin{figure*}[t!]
\centering
\includegraphics[width=0.995\textwidth]{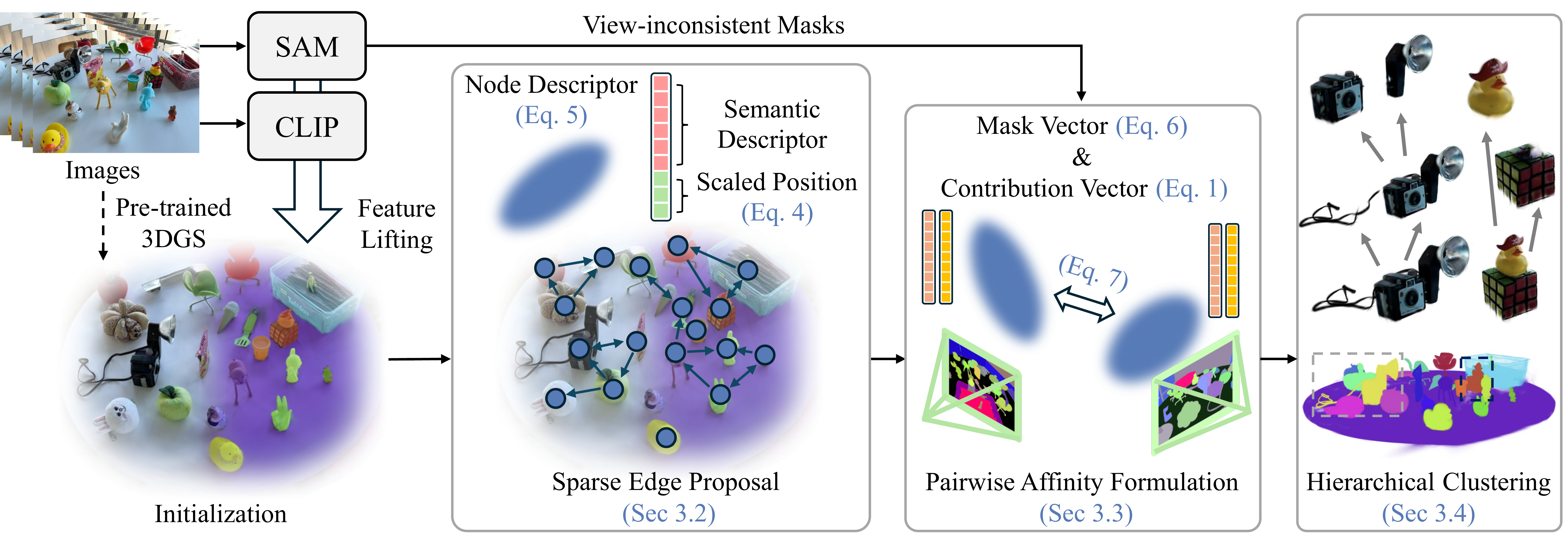}
    \caption{Overview of {\Ours}. 
    Our method begins with a pretrained 3DGS~\cite{3dgs_2023} and initializes its features by lifting SAM~\cite{sam_2023} masked CLIP~\cite{clip_2021} features from each view. Using semantic descriptors and positional cues, we construct low-dimensional node descriptors to propose sparse edge candidates.
    Precise pairwise affinities are then computed only on these candidates to build the relation graph efficiently. Finally, our {\tree} algorithm builds the hierarchical cluster tree.
    }
    \label{fig:overview}
\end{figure*}

%% file: sec/2_related.tex
\section{Related Works}
\label{sec:related}
\paragraph{\textbf{Language-embedded Radiance Fields}}
Prior work has embedded 2D foundation-model features~\cite{clip_2021,sam_2023,dino_2021,lseg_2022} into implicit Neural radiance fields (NeRF~\cite{mildenhall2021nerf}) to enable open-vocabulary querying, segmentation, and editing~\cite{liu2023weakly,engelmann2024opennerf,fu2022panoptic,kerr2023lerf,kobayashi2022decomposing,siddiqui2023panoptic,tschernezki2022neural,kundu2022panoptic,zhang2023nerflets}. 
More recently, explicit 3D Gaussian Splatting (3DGS~\cite{3dgs_2023}) has emerged as a practical substrate for 3D scene understanding due to its efficient optimization and real-time rendering. 
In this context, \emph{language-based} 3DGS methods~\cite{feature3dgs_2024, langsplat_2024, goi_2024, legaussian_2024, fmgs_2025, vlgs_2024} distill knowledge from 2D vision-language models by optimizing rendered features.

Subsequent works~\cite{occams_2024,drsplat_2025,cf3_2025,ludvig_2024} move toward more 3D-aware lifting by leveraging Gaussian visibility and view-dependent rendering contributions: center-based lifting provides a lightweight approximation~\cite{occams_2024}, while contribution-weighted lifting aggregates evidence from all contributing pixels for higher fidelity~\cite{drsplat_2025,ludvig_2024,cf3_2025}.

Unlike prior language-based methods that directly use per-Gaussian semantic features for open-vocabulary 3D Gaussian segmentation, {\Ours} uses them only to form semantic descriptors.
Combined with locality cues, these descriptors are used to propose sparse edge candidates.
Segmentation is then derived by explicitly modeling relations between Gaussians, which leads to more coherent results.

\paragraph{\textbf{Instance-feature-based Methods}}
Instance-feature-based methods can be grouped by how they handle view-inconsistent multi-view 2D masks~\cite{sam_2023}.
In particular, they adopt either implicit or explicit strategies. 

Implicit approaches~\cite{opengaussian_2024,instancegaussian_2025,laga_2025,saga_2025,supergseg_2024,clickguassian_2024,votesplat_2025,cos3d_2025} use view-inconsistent masks in contrastive learning to optimize instance-feature. Works like OpenGaussian~\cite{opengaussian_2024}, InstanceGaussian~\cite{instancegaussian_2025}, LaGa~\cite{laga_2025}, and SuperGSeg~\cite{supergseg_2024} differ in ID assignment, using codebooks, graph-based merging, post-hoc clustering, or Super-Gaussians. Other methods~\cite{clickguassian_2024,saga_2025,votesplat_2025} incorporate auxiliary cues such as semantics or physical priors, or adopt hough-voting formulations to improve clustering under view inconsistency.
Also, COS3D~\cite{cos3d_2025} leverages learned instance features to support bidirectional mapping between instance and language representations.

Explicit approaches~\cite{gaussiangrouping_2024,objectgs_2025,gaga_2024,trace3d_2025} resolve view inconsistency by associating masks across views. Gaussian Grouping~\cite{gaussiangrouping_2024} and ObjectGS~\cite{objectgs_2025} leverage external trackers~\cite{sam2_2024,deva_2023} to unify mask identities before performing instance-feature learning on 3DGS. However, relying on external tracking for cross-view association can be brittle to initialization and to large viewpoint changes. Beyond tracker-based pipelines~\cite{gaussiangrouping_2024,objectgs_2025}, some methods~\cite{gaga_2024,trace3d_2025} aim to incorporate 3D-aware mask association during training to improve cross-view consistency.

Furthermore, training-free approaches~\cite{lbg_2025,thgs_2025, lightsplat_2026, extrinsplat_2026, relags_2026} have been explored, often focusing on computational speed. THGS~\cite{thgs_2025} adapts point cloud (PC) algorithms by treating Gaussians as simple points \texttt{([pos, color, normal])} from 2DGS~\cite{2dgs_2024} and applying superpoint clustering~\cite{cutpursuit_2017, sai3d_2024}. While this abstraction enables efficient grouping, it does not account for noisy Gaussians near object boundaries, making it less suitable for clean 3D segmentation.

In contrast, {\Ours} is also training-free, but explicitly leverages 3DGS-specific properties such as view contribution and visibility. This allows {\Ours} to produce clean 3D segmentation with high visual quality. In addition, our fast variant runs over $2\times$ faster than THGS while achieving superior performance.



%% file: sec/3_method.tex
\section{Method}
\subsection{Preliminary}
\label{sec:preliminary}
3DGS scene $\mathcal{S}=\{g_i | i=1, \cdots, N\}$ is represented with $N$ Gaussians, where each Gaussian has a center coordinate $\boldsymbol{\mu}\in \mathbb{R}^3$, a covariance matrix ${\Sigma}\in \mathbb{R}^{3\times 3}$, an opacity $\alpha\in\mathbb{R}_+$.
Color ${C}$ at each image pixel is rendered via alpha-blending of spherical-harmonic features $\boldsymbol{c}_i$, accounting for depth ordering with respect to the viewpoint.
\begin{equation}
    {C}  = \sum_{i=1}^N \boldsymbol{c}_i\alpha_i \prod^{i-1}_{j=1}(1-\alpha_j) =\sum_{i=1}^N \boldsymbol{c}_i \alpha_i T_i =\sum_{i=1}^N \boldsymbol{c}_iw_i, 
\label{eq:3dgs_color}
\end{equation}
where $T_i\in\mathbb{R}_+$ is transmittance. We denote $w_i$ as the \emph{weight of the corresponding Gaussian} contributing to each pixel.

\subsection{Sparse Edge Proposal}
\label{sec:edge_con}

For 3D Gaussian segmentation, we construct a relation graph where nodes are a Gaussian and edge weights represent pairwise affinity.
The pairwise affinity between Gaussians is formulated from multi-view mask evidence, weighted by their view contribution scores.
However, evaluating pairwise relations for all Gaussian pairs in a scene is inefficient.
We therefore adopt a two-stage strategy.
A lightweight node descriptor is first used to coarsely propose sparse candidate edges (\cref{sec:edge_con}). 
Precise pairwise affinities are then computed only for these candidates (\cref{sec:edge_weight}).

We first formulate a node descriptor $\mathbf{z}_i$ for each Gaussian $g_i$ to encapsulate its semantic and geometric characteristics, defined as the concatenation of a semantic descriptor $\mathbf{f}_i^{\text{sem}}$ and a position vector $\boldsymbol{\mu}_i$.
To construct the $\mathbf{f}_i^{\text{sem}}$, we first initialize per-Gaussian semantic features by training-free lifting of SAM-masked CLIP~\cite{sam_2023,clip_2021} features from each view. We use contribution-weighted lifting~\cite{ludvig_2024, drsplat_2025, cf3_2025} for a \textbf{\Ours} initialization, and a center-based variant~\cite{occams_2024} for a \textbf{{\Ours}-Fast}. 
We then project these high-dimensional features into a 6D space using PCA~\cite{pca}, retaining the top 6 components to inject lightweight semantic cues into the node descriptors.
To balance the different scales of $\mathbf{f}_i^{\text{sem}} \in \mathbb{R}^6$ and $\boldsymbol{\mu}_i \in \mathbb{R}^3$ before concatenation, we apply a ratio-based normalization. We independently compute the average distance to the $k$-nearest neighbors within each space:
\begin{equation}
    \bar{d}_{\text{sem}} = \frac{1}{N} \sum_{i=1}^{N} \left( \frac{1}{k} \sum_{j \in \mathcal{N}_{k}(\mathbf{f}_i^{\text{sem}})} \|\mathbf{f}_i^{\text{sem}} - \mathbf{f}_j^{\text{sem}}\|_2 \right),
\end{equation}
\begin{equation}
    \bar{d}_{\text{pos}} = \frac{1}{N} \sum_{i=1}^{N} \left( \frac{1}{k} \sum_{j \in \mathcal{N}_{k}(\boldsymbol{\mu}_i)} \|\boldsymbol{\mu}_i - \boldsymbol{\mu}_j\|_2 \right),
\end{equation}
where $\mathcal{N}_{k}(\cdot)$ denotes the set of indices for the $k$-NN of a given vector. A scaling factor $\lambda$ is then derived from the ratio of these average distances:
\begin{equation}
    \lambda = \frac{\bar{d}_{\text{sem}}}{\bar{d}_{\text{pos}}}.
\end{equation}
Finally, we form the unified node descriptor $\mathbf{z}_i \in \mathbb{R}^9$:
\begin{equation}
\mathbf{z}_i = [\mathbf{f}_i^{\text{sem}}, \lambda \times \boldsymbol{\mu}_i].
\end{equation}


We perform $k$-NN search on the lightweight node descriptors to propose sparse edge candidates.
We construct these edges as a directed graph, where each node connects to its $k$ nearest neighbors (\eg, if Gaussian $g_j$ is among the $k$ nearest neighbors of $g_i$, we add an edge $i \rightarrow j$).
This digraph design preserves consistent local connectivity per node, which stabilizes the subsequent hierarchical clustering.

\subsection{Pairwise Affinity Formulation}
\label{sec:edge_weight}
We formulate instance decomposition as \emph{pairwise relation estimation} on a sparse edge candidates. Given the digraph structure from \cref{sec:edge_con}, we compute an edge weight $e_{ij}$ by accumulating \emph{pairwise evidence across views}. Concretely, we combine multi-view mask observations with each Gaussian's view-dependent contribution. This yields an evidence-weighted affinity score for each edge. A higher score indicates that the two Gaussians are more likely to belong to the same instance.

For each Gaussian $g_i$, we extract two fundamental pieces of information: a discrete mask index vector and a view contribution weight vector. Let $\mathbf{M}_v \in \mathbb{Z}^{H_v \times W_v}$ be the mask map for view $v$. We define two special index values: $\mathcal{I}_{\text{noise}}$ for uncertain or boundary regions, and $\mathcal{I}_{\text{inv}}$ for regions where the Gaussian is invisible (\eg, outside the field of view or occluded).

To robustly handle ambiguities at object boundaries, a Gaussian's mask index must be carefully determined. As shown in \cref{fig:index}, a Gaussian's projected center may fall on a different instance than the area where it contributes most of its color.
We therefore sample the mask index from two distinct locations for each Gaussian $g_i$ in view $v$. The contribution weight $\omega_{iv}$ for Gaussian $g_i$ in view $v$, as derived from \cref{eq:3dgs_color}, is used with this mask index $m_{iv}$.

\noindent\textbf{Center-based Index ($m_{iv}^{\text{center}}$):} The mask index at the projected center of the Gaussian, $\mathbf{p}_{iv}^{\text{center}}$.

\noindent\textbf{Max-based Index ($m_{iv}^{\text{max}}$):} The mask index at the pixel coordinate $\mathbf{p}_{iv}^{\text{max}}$ where the Gaussian's contribution weight is maximal, \ie, $\mathbf{p}_{iv}^{\text{max}} = \arg\max_{\mathbf{p}} w_{i,\mathbf{p}}^{(v)}$.
\input{figs/index}

\begin{equation}
\label{eq:robust_mask}
m_{iv} = 
\begin{cases} 
\mathcal{I}_{\text{noise}} & \text{if } m_{iv}^{\text{center}} \neq m_{iv}^{\text{max}} \\
m_{iv}^{\text{max}} & \text{otherwise} 
\end{cases}
\end{equation}

These two indices are reconciled into a single, mask index $m_{iv}$ from \cref{eq:robust_mask}. Our logic prioritizes the Max-weight Index ($m_{iv}^{\text{max}}$) as the default, as it indicates where the Gaussian's visual properties (color, semantics) are expressed. The Center-based Index ($m_{iv}^{\text{center}}$) is used as a cross-check to detect ambiguities. Specifically, if the center and max-weight indices differ, we consider the Gaussian to be on a boundary and mark it as noise ($\mathcal{I}_{\text{noise}}$). 

With the mask index vector $\mathbf{m}_i, \mathbf{m}_j \in \mathbb{Z}^V$ and contribution vector $\boldsymbol{\omega}_i, \boldsymbol{\omega}_j \in \mathbb{R}^V$ defined for two Gaussians $g_i$ and $g_j$, we can formulate the edge weight $e_{ij}$. First, we define a set of \textit{comparable} views, $J_{ij} \subseteq \{1, ..., V\}$, for the pair $(i,j)$. A view $v$ is considered comparable if it provides reliable evidence for both Gaussians, excluding pairs that are invisible ($\mathcal{I}_{\text{inv}}$) or consist purely of noise ($\mathcal{I}_{\text{noise}}$).

The edge weight $e_{ij}$ is formulated as the weighted ratio of co-occurrence as follows:
\begin{equation} 
\label{eq:edge_weight}
e_{ij} = \frac{\sum_{v \in J_{ij}} \mathbb{I}(m_{iv} = m_{jv}) \cdot (\omega_{iv} \cdot \omega_{jv})}{\sum_{v \in J_{ij}} (\omega_{iv} \cdot \omega_{jv})},
\end{equation}
where we define the \textit{pairwise evidence weight} of a pair $(i, j)$ in a view $v$ as the product of their contributions, $\omega_{iv} \cdot \omega_{jv}$. Using the product acts as a strict logical AND gate: the joint contribution is high \textit{only if} both Gaussians are strongly visible in that view. This is a robust measure that penalizes pairs where one Gaussian is dominant and the other is negligible.
The total edge weight $e_{ij}$ is then the ratio of this joint contribution from views where the Gaussians agree (share the same mask index) divided by the total joint contribution from all comparable views.

Here, $\mathbb{I}(\cdot)$ is the indicator function, which is 1 if the condition is true and 0 otherwise. The condition $m_{iv} = m_{jv}$ implicitly excludes cases where the common index is $\mathcal{I}_{\text{noise}}$ due to the definition of $J_{ij}$. The resulting weight $e_{ij} \in [0, 1]$ represents the affinity, where a higher value indicates a stronger probability that $g_i$ and $g_j$ belong to the same instance. This calculation is performed for all $N \times k$ directed edges in the sparse relation graph.

\input{figs/tree}
\subsection{Hierarchical Clustering}
\label{sec:tree_dbscan}
To handle multi-granular queries, a flexible structure is needed rather than a flat, one-layer scene split. Given the precomputed relation graph with edge weights $e_{ij}$, our goal is to construct a cohesive hierarchical cluster tree.
A na\"ive approach is to run DBSCAN multiple times with varying thresholds, which produces a set of clusters across levels.
However, repeated independent clustering does not make parent-child relations between levels.
Therefore, it cannot distinguish meaningful structural splits from spurious fragmentation and often retains redundant clusters across levels, which is also observed in previous hierarchical pipelines~\cite{lbg_2025,thgs_2025,laga_2025}.

To address this, we introduce {\tree}, which builds \emph{a parent-child-aware cluster tree} from a weighted digraph.
{\tree} enforces inter-level relational constraints to prune redundant and spurious clusters, yielding a single cohesive hierarchy of meaningful clusters.

\input{table/tree_dbscan2}
As described in \cref{sec:edge_con}, our relation graph is a sparse digraph where each node has an out-degree of $k$, connecting to its $k$-nearest neighbors.
We modify DBSCAN for this setting.
Unlike DBSCAN, our method uses an affinity threshold $\theta$ instead of a distance radius $\epsilon$ and relies only on out-edges.
This design is motivated by graph asymmetry: in-degrees can vary widely, while each node always has exactly $k$ out-edges.
This fixed local connectivity improves stability across diverse scenes. Also,  we set \texttt{min\_samples} ($m_p$) to $k$ to require all out-edge weights to exceed $\theta$ to form core points. This helps improve segmentation near object boundaries.

{\tree} operates over $L$ levels using a monotonically increasing sequence of affinity thresholds $\Theta=\{\theta_1,\theta_2,\ldots,\theta_L\}$. In our experiments, we use a shared threshold schedule across scenes. The procedure, summarized in \cref{alg:tree_dbscan_compact}, is as follows:

\input{table/LERF}
\input{figs/qual_lerf}

\noindent\textbf{Initialization (Level 1):} Initially, we run DBSCAN on the digraph $G$ with the minimum affinity threshold $\theta_1$. The resulting clusters in $\mathcal{C}_1$ constitute the root level of our tree. (line 4-5 in \cref{alg:tree_dbscan_compact}) 

\noindent\textbf{Hierarchical Splitting (Level $l>1$):} For each valid parent cluster $C_p \in \mathcal{C}_{l-1}$ from the previous level, we run DBSCAN exclusively on its subgraph using the next threshold, $\theta_l$ (line 8-9 in \cref{alg:tree_dbscan_compact}). By ensuring that clusters are only partitioned within their parent boundaries, this containment relationship serves as the foundation for the subsequent pruning process.

\noindent\textbf{Hierarchical Pruning:} As illustrated in \cref{fig:tree}, we apply two validation steps after each potential split (line 10-16 in \cref{alg:tree_dbscan_compact}):
\begin{itemize}
    \item \textbf{Redundancy Filtering:} 
    If a parent cluster $C_p$ fails to split into at least two children ($|\mathcal{C}_\text{child}| < 2$), it indicates that the clustering at this level merely replicates the parent structure. To prevent the creation of redundant nodes, $\mathcal{C}_{\text{child}}$ is not added to the tree $\mathcal{T}$ in this case. Instead, $C_p$ is carried forward to $\mathcal{C}_l$ to be re-evaluated at the next level ${l+1}$ where a more meaningful partition may occur.
    \item \textbf{Spurious Cluster Filtering:} While partitioning naturally leads to the conversion of some Gaussians into noise, an excessive loss of cluster mass indicates that the split is spurious. To quantify this effect, we compute the erosion rate $\rho$ whenever a split occurs ($|\mathcal{C}_\text{child}| \ge 2$):
    \begin{equation}
    \label{eq:erosion_rate}
    \rho = \frac{\sum_{C' \in \mathcal{C}_\text{child}}|C'|}{|C_p|}.
    \end{equation}
     If $\rho$ falls below $\rho_{\min}$ (set to 0.5 by default), the split is deemed unreliable and the clusters in $\mathcal{C}_{\text{child}}$ are pruned (\ie, excluded from both $\mathcal{C}_l$ and $\mathcal{T}$).
\end{itemize}


\noindent\textbf{Semantic refinement for open-vocabulary querying.}
After hierarchical clustering, we refine cluster-level semantic features through a lightweight 3D-2D association based on rasterized cluster maps and masks, avoiding instance-feature similarity filtering used in some prior pipelines (\eg, OpenGaussian~\cite{opengaussian_2024}). See supplementary material for full details.

%% file: figs/index.tex
\begin{figure}[t!]
\centering
\includegraphics[width=1.0\columnwidth]{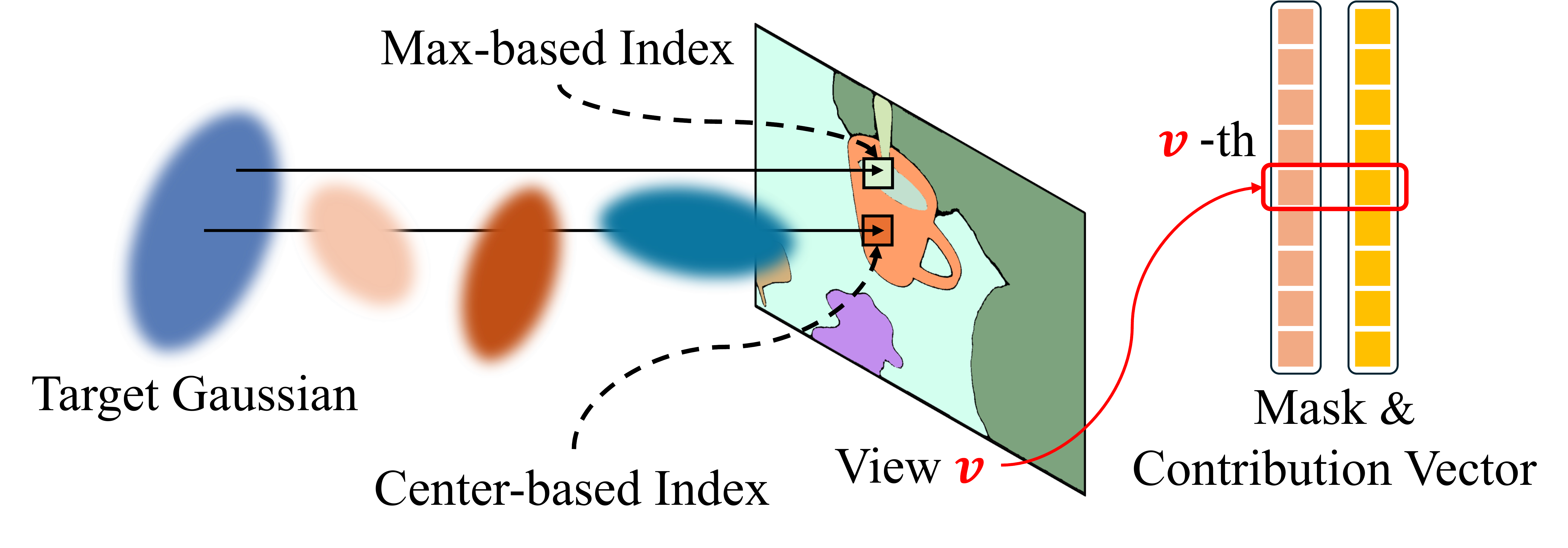}
    \caption{Illustration of the two sampling locations for robust mask index retrieval. The Center-based Index ($m^{\text{center}}$) samples at the projected center, while the Max-based Index ($m^{\text{max}}$) samples where the Gaussian's contribution is highest. This handles cases where the center lies on a boundary.}
    \label{fig:index}
\vspace{-3mm}
\end{figure}

%% file: figs/tree.tex
\begin{figure}[t!]
\centering
\includegraphics[width=1.0\columnwidth]{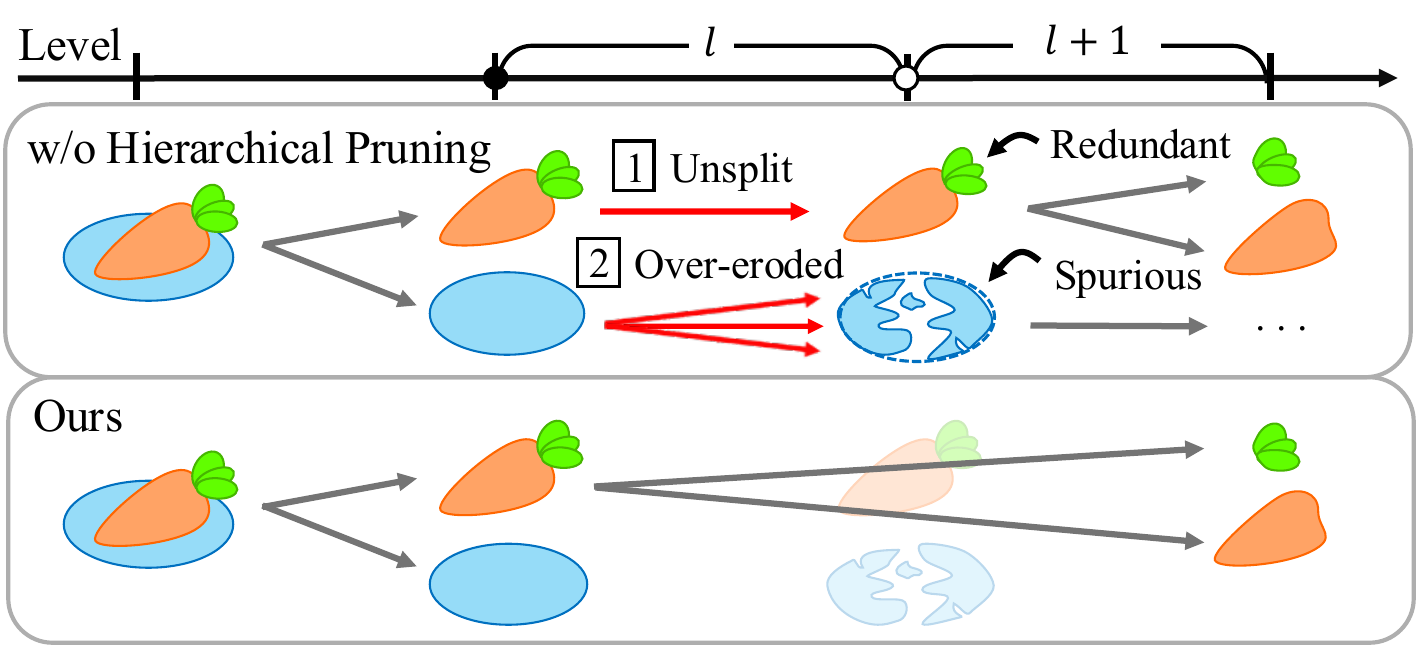}
    \caption{Illustration of the Hierarchical Pruning steps in \tree. \fbox{$1$} and \fbox{$2$} present two filtering cases: Redundancy Filtering and Spurious Cluster Filtering, respectively.
    These procedures prevent the formation of uninformative clusters (shown as faded) at each level, enabling the construction of an efficient cluster tree.}
    \label{fig:tree}
\vspace{-3mm}
\end{figure}

%% file: table/tree_dbscan2.tex
\begin{algorithm}[t!]
\caption{\tree}
\label{alg:tree_dbscan_compact}
\small
\begin{algorithmic}[1]
\State \textbf{Input:} Digraph $G=(V, E, W)$, Thresholds $\Theta=\{\theta_1,\theta_2,\dots,\theta_L\}$, $m_p$, $\rho_e$.
\State \textbf{Output:} Cluster tree $\mathcal{T}$.

\State \textbf{Initialize:}
\State $\mathcal{C}_1 \leftarrow \text{DBSCAN}(G, \theta_1, m_p)$
\State $\mathcal{T} \leftarrow \text{InitializeTree}(\mathcal{C}_1)$

\For{$l = 2$ \textbf{to} $L$} 
    \State $\mathcal{C}_l \leftarrow \emptyset$
    \For{\textbf{each} parent cluster $C_{p} \in \mathcal{C}_{l-1}$}
        \State $\mathcal{C}_{\text{child}} \leftarrow \text{DBSCAN}(G[C_p], \theta_l, m_p)$
        
    \If{$|\mathcal{C}_{\text{child}}| < 2$}
        \Comment{1. Check Redundancy}
        \State $\mathcal{C}_l \leftarrow \mathcal{C}_l \cup \{C_p\}$
    \Else
        \State $\rho \leftarrow \left( \sum_{C' \in \mathcal{C}_{\text{child}}} |C'| \right) / |C_p|$
        \If{$\rho \ge \rho_e$}
            \Comment{2. Check Erosion rate}
            \State Add each $C' \in \mathcal{C}_{\text{child}}$ as a child of $C_p$ in $\mathcal{T}$
            \State $\mathcal{C}_l \leftarrow \mathcal{C}_l \cup \mathcal{C}_{\text{child}}$
        \EndIf
    \EndIf
    \EndFor
\EndFor
\State \Return $\mathcal{T}$
\end{algorithmic}
\end{algorithm}

%% file: table/LERF.tex
\begin{table*}[t!]
\caption{\textbf{Evaluation on LERF~\cite{lerf_2023} dataset.} We compare language-based methods (LangSplat~\cite{langsplat_2024},Occam's LGS~\cite{occams_2024}, Dr.Splat~\cite{drsplat_2025}), instance-feature-based methods (OpenGaussian~\cite{opengaussian_2024}, InstanceGaussian~\cite{instancegaussian_2025}, COS3D~\cite{cos3d_2025} LaGa~\cite{laga_2025}), and a point-based method (THGS~\cite{thgs_2025}).
 {\Ours}-Fast denotes our fast version, which utilizes center-based lifting~\cite{occams_2024} for feature initialization (\cref{sec:edge_con}). The two versions differ only in this initialization method. The high-fidelity version (unmarked) is our default. Accurate is measured by mAcc$@$0.50.}
\centering
\resizebox{0.9\textwidth}{!}{
\setlength{\tabcolsep}{6pt}
\begin{tabular}{l|ccccc|ccccc|r}
\toprule[1.2pt]
& 
\multicolumn{5}{c|}{\textbf{mIoU$\uparrow$}} &
\multicolumn{5}{c|}{\textbf{mAcc$\uparrow$}} &
\textbf{Time$\downarrow$} \\ 
&Figurines & Ramen &Teatime &Waldo & \textbf{Mean} 
&Figurines & Ramen &Teatime &Waldo & \textbf{Mean} & (min:sec)\\ 
\midrule
LangSplat~\cite{langsplat_2024} &24.2 &19.7 &36.7 &28.9 &27.4 &8.9 &9.4 &29.9 &19.7 &17.0 &88:04 \\
Occam's LGS~\cite{occams_2024} &52.0 &25.7 &60.8 &46.9 &46.4
&57.1 &22.5 &\cellcolor{colort}76.3 &40.9 &49.2 &\cellcolor{colorf}0:31\\
Dr.Splat~\cite{drsplat_2025} &51.8	&30.5	&59.5	&45.5	&46.8 
&55.4	&21.1	&72.9	&45.5	&48.7  &3:37\\
\midrule
OpenGaussian~\cite{opengaussian_2024} &57.6	&21.2	&57.6	&34.9	&42.8
&\cellcolor{colort}66.1	&12.7	&67.8	&36.4	&45.7 &48:42\\
InstanceGaussian~\cite{instancegaussian_2025} &48.8	&29.3	&54.6	&42.3	&43.8
&58.9	&\cellcolor{colort}32.4	&64.4	&45.5	&50.3 &213:04 \\

COS3D~\cite{cos3d_2025} &\cellcolor{colort}59.6	&30.7	&\cellcolor{colors}66.7	&39.1	&49.0
&\cellcolor{colort}66.1	&25.4	&\cellcolor{colorf}83.1	&36.4	&52.7 &50:18 \\
LaGa~\cite{laga_2025} &52.8	&\cellcolor{colors}40.5	&63.5	&\cellcolor{colort}52.0	&\cellcolor{colort}52.2
&55.4	&\cellcolor{colors}39.4	&71.2	&\cellcolor{colort}54.5	&\cellcolor{colort}55.1 & 74:47 \\
\midrule
THGS~\cite{thgs_2025} &41.3	&27.8	&56.6	&44.2	&42.5 
&39.3	&29.6	&67.8	&36.4	&43.3
&\cellcolor{colort}1:59\\


{\Ours}-Fast (Ours) &\cellcolor{colors}70.0	&\cellcolor{colort}32.7	&\cellcolor{colorf}66.9	&\cellcolor{colors}57.5	&\cellcolor{colors}56.8
&\cellcolor{colorf}80.4	&\cellcolor{colort}32.4	&\cellcolor{colors}78.0	&\cellcolor{colors}63.6 &\cellcolor{colors}63.6  &\cellcolor{colors}0:58\\
{\Ours} (Ours) &\cellcolor{colorf}70.1	&\cellcolor{colorf}44.0	&\cellcolor{colort}66.5	&\cellcolor{colorf}60.8	&\cellcolor{colorf}60.4
&\cellcolor{colors}78.6	&\cellcolor{colorf}47.9	&\cellcolor{colors}78.0	&\cellcolor{colorf}68.2	&\cellcolor{colorf}68.2 &3:02\\
\bottomrule[1.2pt]
\end{tabular}
}
\label{tab:lerf}
\end{table*}

%% file: figs/qual_lerf.tex
\begin{figure*}[t!]
\centering
\vspace{4mm}
\includegraphics[width=0.9\textwidth]{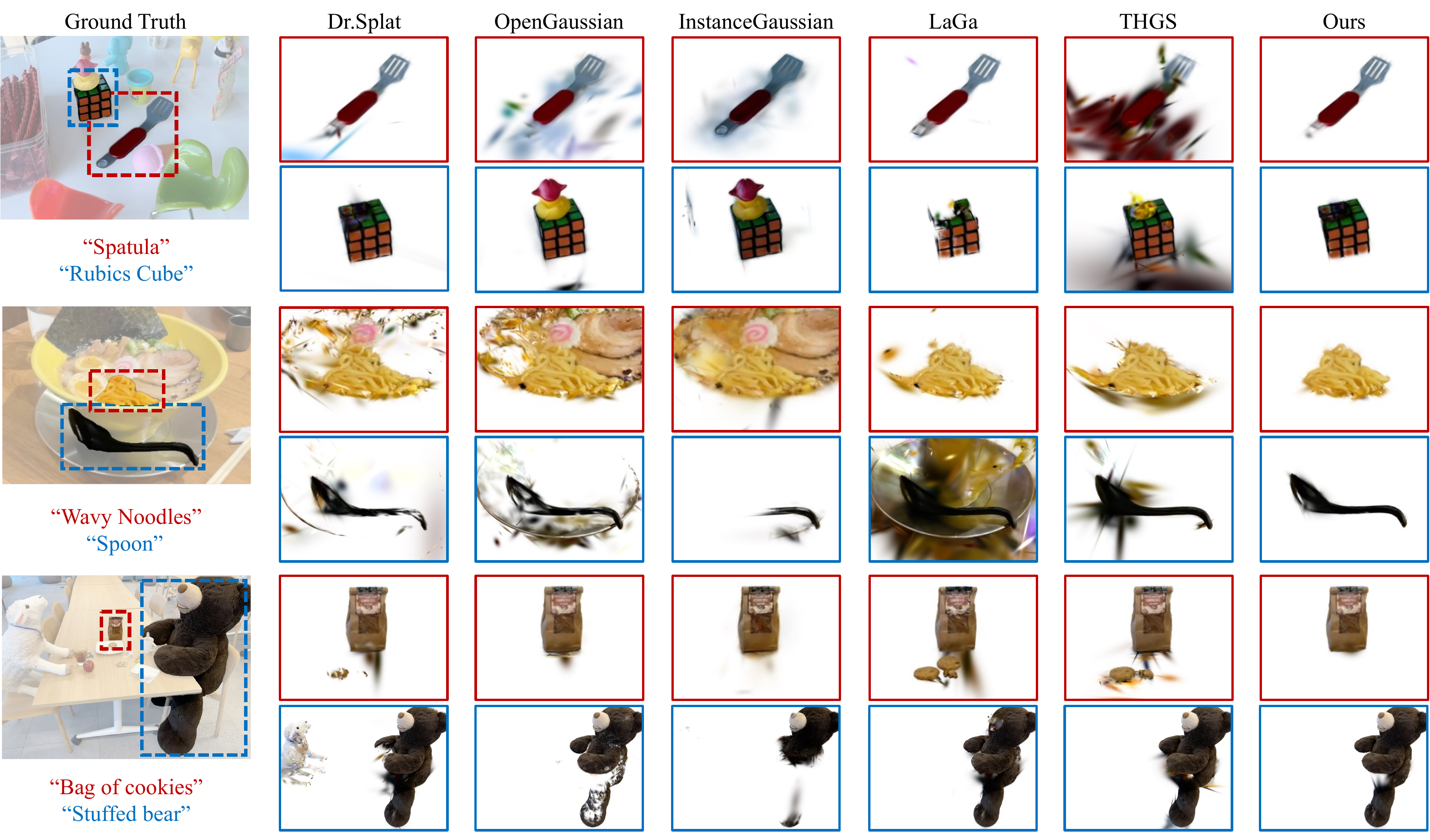}
    \caption{\textbf{Qualitative comparison.} We visualize open-vocabulary object selection results on LERF~\cite{lerf_2023} dataset. 
    Compared to language-based~\cite{drsplat_2025} and instance-feature-based~\cite{opengaussian_2024,instancegaussian_2025,laga_2025} baselines, PairGS produces cleaner 3D Gaussian segmentations for objects in contact with others, such as a spoon or a spatula.
    It also yields coherent instance separation for larger objects. In particular, THGS~\cite{thgs_2025} is point-based and considers only Gaussian centers, which often leads to noisy Gaussians in contact regions and along object boundaries.}
    \label{fig:qual_lerf}
\end{figure*}

%% file: sec/4_experiments.tex
\input{table/Scannet_sem}
\section{Experiments}
\label{sec:experiments}
To evaluate our method, we conduct comparative experiments against a set of state-of-the-art methods carefully selected to represent each major category, including language-based methods ~\cite{langsplat_2024,occams_2024,drsplat_2025}, instance-feature-based methods~\cite{opengaussian_2024,instancegaussian_2025,laga_2025,cos3d_2025} and point-based method~\cite{thgs_2025}. We report results for two variants of our pipeline: {\Ours} using contribution-weighted intialization, and {\Ours}-Fast using center-based intialization.
We evaluate segmentation performance (\eg, mIoU and accuracy) and efficiency measured by total runtime, with all experiments run on a single NVIDIA RTX 6000 Ada GPU. For fair comparison, all experiments follow OpenGaussian’s~\cite{opengaussian_2024} 3D evaluation protocol. Baselines use the hyperparameter settings recommended in their original papers, while our method uses a shared hyperparameter setting across all experiments.


\subsection{Open-Vocabulary 3D Object Selection}
\label{lerf_eval}
\noindent\textbf{Dataset.}
We evaluate open-vocabulary object selection on the LERF-OVS dataset~\cite{lerf_2023}. For evaluation, each method selects target Gaussians for a given text query based on its specific metric (\eg, relevancy score~\cite{langsplat_2024} or cosine similarity). We then render the selected Gaussians to compare against the ground-truth (GT) mask, reporting mIoU and accuracy (Acc.).

\noindent\textbf{Baselines.}
We compare against language-based methods (LangSplat~\cite{langsplat_2024}, Occam's LGS~\cite{occams_2024}, Dr.Splat~\cite{drsplat_2025}), instance-feature methods (OpenGaussian~\cite{opengaussian_2024}, InstanceGaussian~\cite{instancegaussian_2025}, COS3D~\cite{cos3d_2025}, LaGa~\cite{laga_2025}), and point-based baseline (THGS~\cite{thgs_2025}). Language-based methods~\cite{langsplat_2024,occams_2024,drsplat_2025} segment Gaussians by thresholding per-Gaussian query scores, so we carefully select the threshold via a sweep to maximize the overall performance.
Most methods utilize the same pre-trained 3DGS~\cite{3dgs_2023} model for fair comparison, with the exception of InstanceGaussian~\cite{instancegaussian_2025} (Scaffold-GS~\cite{scaffold_2024}) and THGS~\cite{thgs_2025} (2DGS~\cite{2dgs_2024}). 
We note that some baselines~\cite{opengaussian_2024,langsplat_2024,laga_2025,thgs_2025} involve specific configuration. 
OpenGaussian~\cite{opengaussian_2024}, for example, requires a different number of centroids to be specified for each scene. For LangSplat~\cite{langsplat_2024}, LaGa~\cite{laga_2025} and THGS~\cite{thgs_2025}, the methods use all 3 levels of SAM~\cite{sam_2023} masks proposed in LangSplat~\cite{langsplat_2024}. In contrast, our method and all other baselines~\cite{drsplat_2025,occams_2024,opengaussian_2024,instancegaussian_2025,cos3d_2025} use only the single, whole-level mask. 

As shown in \cref{tab:lerf}, {\Ours} achieves the highest mIoU and accuracy, outperforming even the 3-level LaGa~\cite{laga_2025} implementation while using only a single mask level. Furthermore, {\Ours}-Fast demonstrates superior efficiency, running faster than the efficiency-focused THGS~\cite{thgs_2025} while maintaining higher performance.

Qualitatively, \Cref{fig:qual_lerf} shows that our method produces cleaner and more fine-grained object boundaries than both THGS~\cite{thgs_2025} and instance-feature-based methods~\cite{opengaussian_2024,instancegaussian_2025,laga_2025}.
By constructing a cohesive hierarchical cluster tree with TreeDBSCAN, our method supports accurate object selection at multiple granularities, in contrast to instance-feature methods~\cite{opengaussian_2024,instancegaussian_2025,laga_2025}. Moreover, compared to Dr.Splat~\cite{drsplat_2025}, our cohesive clusters reduce per-Gaussian noise.

\subsection{Open-Vocabulary Scene Understanding}
\label{scannet_eval}
\noindent\textbf{Dataset.}
We evaluate open-vocabulary scene understanding on the ScanNet dataset~\cite{scannet_2017}. To ensure a fair point-wise comparison against ground truth, we freeze Gaussian coordinates during optimization to prevent densification. We report metrics for both open-vocabulary semantic segmentation and class-agnostic instance segmentation.

\noindent\textbf{Baselines.}
For semantic segmentation, we use the same baselines as in Sec.~4.1. 
For instance segmentation, we include state-of-the-art approaches, such as Gaussian Grouping~\cite{gaussiangrouping_2024} and ObjectGS~\cite{objectgs_2025}, which use a video tracker~\cite{deva_2023} for mask association. Notably, InstanceGaussian~\cite{instancegaussian_2025} utilizes an additional MLP to predict point labels during scene training.

\input{table/Scannet_Ins}

\Cref{tab:scannet_sem} shows the open-vocabulary semantic segmentation results. Our method achieves the highest mIoU, outperforming even InstanceGaussian~\cite{instancegaussian_2025} despite its use of an auxiliary MLP for labeling. This is visually corroborated in \cref{fig:qual_scannet}, where our method demonstrates accurate segmentation with clean boundaries compared to baselines.

To evaluate the clustering quality itself, we also perform the class-agnostic instance segmentation proposed in InstanceGaussian~\cite{instancegaussian_2025}, as shown in \cref{tab:scannet_ins}. Tracker-based methods~\cite{gaussiangrouping_2024, objectgs_2025} show limited performance due to inaccurate associations by off-the-shelf trackers~\cite{deva_2023} on ScanNet. In contrast, {\Ours} achieves stronger class-agnostic instance segmentation results, outperforming instance-feature-based methods~\cite{opengaussian_2024,laga_2025,instancegaussian_2025}.

\input{figs/qual_scannet}

\input{table/only_ablation}
\subsection{Ablation}
We conduct ablation studies to assess the impact of key design choices in our pipeline.
As shown in \cref{tab:ablation}, we examine individual components within sparse edge candidate proposal (\cref{sec:edge_con}), pairwise affinity formulation (\cref{sec:edge_weight}) and hierarchical clustering (\cref{sec:tree_dbscan}).

For graph construction, the results show that the positional scaling factor ($\lambda$) is important for balancing the semantic descriptor and the position vector. 
We further ablate graph directionality by converting the directed graph into an undirected graph. This degrades performance by altering the neighborhood structure used for clustering. Even with mutual $k$-NN, the graph does not fully recover the original connectivity pattern, resulting in lower performance.


For pairwise affinity formulation, we demonstrate the importance of our Gaussian-aware design. Removing the view contribution vector ($\omega_{iv}$) or using only the center-based index ($m_{iv}^{\textbf{center}}$) makes 3D Gaussians behave similarly to points. As a result, the pairwise affinity in \cref{eq:edge_weight} cannot reflect each Gaussian's view contribution, and noisy Gaussians cannot be handled by \cref{eq:robust_mask}, leading to a performance drop.

Finally, our hierarchical clustering ablation highlights the need for a hierarchy beyond single-level clustering used in OpenGaussian~\cite{opengaussian_2024} and InstanceGaussian~\cite{instancegaussian_2025}.
Accurate 3D Gaussian segmentation for multi-granular queries, such as \texttt{camera}, \texttt{camera strap}, and \texttt{reflector}, requires a hierarchical representation.
TreeDBSCAN builds a meaningful and coherent cluster tree that supports such multi-granular selection. 


%% file: table/Scannet_sem.tex
\begin{table}[t!]
\caption{Semantic segmentation results on ScanNet~\cite{scannet_2017} dataset.}
    \label{tab:scannet_sem}
    \centering
    \resizebox{1.0\columnwidth}{!}{
    \setlength{\tabcolsep}{2pt}
    \begin{tabular}{l| c c | c c | c c | r}
    \toprule[1.2pt] 
    & 
    \multicolumn{2}{c|}{\textbf{19 classes}} &
    \multicolumn{2}{c|}{\textbf{15 classes}} &
    \multicolumn{2}{c|}{\textbf{10 classes}} &
    \textbf{Time$\downarrow$}\\
    & mIoU$\uparrow$ & mAcc$\uparrow$
    & mIoU$\uparrow$ & mAcc$\uparrow$
    & mIoU$\uparrow$ & mAcc$\uparrow$ & (min:sec)
    \\
    \midrule
    LangSplat~\cite{langsplat_2024} &3.8 &9.1 &5.4 &13.2 &8.4 &22.1 &61:35\\
    Occam's LGS~\cite{occams_2024} &27.6  &39.3 &30.1  &42.8 &36.4  &49.2 &\cellcolor{colorf}0:08\\
    Dr.Splat~\cite{drsplat_2025} & 29.3  &44.7	&33.4  &50.4	&41.3 &58.7  &\cellcolor{colors}0:12\\
    \midrule
    OpenGaussian~\cite{opengaussian_2024} &29.3	 &43.8	&31.8 &47.6	&40.8 &56.8 &19:56\\
    InstanceGaussian~\cite{instancegaussian_2025} &\cellcolor{colort}39.5	&\cellcolor{colort}53.9  &\cellcolor{colort}40.2	&\cellcolor{colort}57.4  &\cellcolor{colort}47.3 &\cellcolor{colort}65.5 &60:59\\
    COS3D~\cite{cos3d_2025} &28.6	&43.7  &34.6	&51.2  &41.9 &58.1 &41:52\\
    LaGa~\cite{laga_2025} &31.7 &48.8	&34.2	&52.4 &41.5 &60.6 &39:34\\
    \midrule
    THGS~\cite{thgs_2025} &33.5 &49.3 &39.3 &56.0 &46.9 &63.9 &0:45\\
    {\Ours}-Fast (Ours) &\cellcolor{colors}40.0 &\cellcolor{colorf}54.8 &\cellcolor{colors}42.5 &\cellcolor{colors}58.5 &\cellcolor{colors}52.5 &\cellcolor{colorf}69.1 &\cellcolor{colort}0:19\\
    {\Ours} (Ours) &\cellcolor{colorf}40.7  &\cellcolor{colors}54.3 &\cellcolor{colorf}44.8 &\cellcolor{colorf}60.2 &\cellcolor{colorf}53.0 &\cellcolor{colors}68.1 &0:36\\
    \bottomrule[1.2pt]
    \end{tabular}}

\end{table}

%% file: table/Scannet_Ins.tex

\begin{table}[t!]
\centering
    \caption{Instance segmentation results on ScanNet~\cite{scannet_2017} dataset.}
    \label{tab:scannet_ins}
    \small
    \resizebox{0.67\columnwidth}{!}{ 
    \setlength{\tabcolsep}{6pt}
    \begin{tabular}{l| c c}
    \toprule[0.8pt] 
    & mIoU$\uparrow$ & mAcc$\uparrow$
    \\
    \midrule
    Gaussian Grouping~\cite{gaussiangrouping_2024} & 19.4 & 4.8\\
    ObjectGS~\cite{objectgs_2025} & 22.7  & 9.8 \\
    \midrule
    OpenGaussian~\cite{opengaussian_2024} & 33.7  & 18.5\\
    InstanceGaussian~\cite{instancegaussian_2025} &\cellcolor{colors}49.9  &\cellcolor{colors}53.9\\
    LaGa~\cite{laga_2025} & 25.2  & 6.0\\
    \midrule
    THGS~\cite{thgs_2025} & 47.6  & 47.0\\
    {\Ours}-Fast (Ours) &\cellcolor{colort}48.9  &\cellcolor{colort}49.4\\
    {\Ours} (Ours) &\cellcolor{colorf}51.9  &\cellcolor{colorf}56.1\\
    \bottomrule[0.8pt]
    \end{tabular}}
\vspace{-3mm}
\end{table}

%% file: figs/qual_scannet.tex
\begin{figure*}[t!]
\centering
\includegraphics[width=0.9\textwidth]{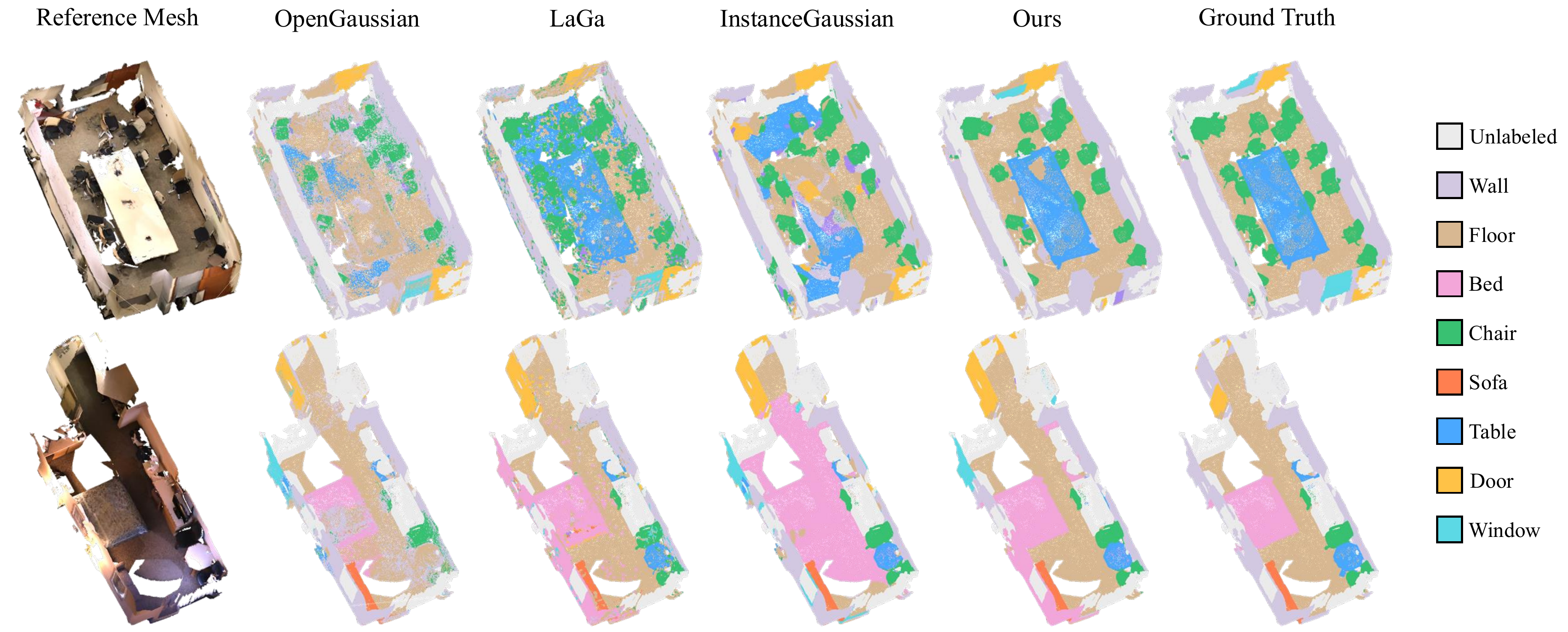}
    \caption{\textbf{Qualitative comparison.} We visualize open-vocabulary semantic segmentation on ScanNet~\cite{scannet_2017} dataset. Our method demonstrates robust performance across diverse queries, accurately segmenting objects at varying scales, such as large desks and smaller chairs.}
    \label{fig:qual_scannet}
\end{figure*}

%% file: table/only_ablation.tex
\begin{table}[t]
  \centering
  \small
  \caption{Ablation study on LERF~\cite{lerf_2023} dataset.}
  \label{tab:ablation}
  \resizebox{0.8\columnwidth}{!}{
  \setlength{\tabcolsep}{8pt}
  \begin{tabular}{l | c c}
  \toprule[1.2pt] 
   & mIoU$\uparrow$ & mAcc$\uparrow$\\
  \midrule
  w/o Scaled Position & 57.5 & 64.0 \\
  Undirected Graph (basic $k$-NN) & 48.7 & 52.3\\
  Undirected Graph (mutual $k$-NN) & 57.6 & 66.2 \\
  \midrule    
  w/o Contribution Vector & 57.4 & 62.7\\
  w/o Max-based-Index & 52.7 & 57.8 \\
  \midrule
  w/o TreeDBSCAN & 54.9 & 60.0\\
  \midrule
  Full (Ours) & \textbf{60.4} & \textbf{68.2}\\
  \bottomrule[1.2pt]
  \end{tabular}
  }
  \vspace{-3mm}
\end{table}

%% file: sec/5_conclusion.tex
\section{Conclusion}

In this paper, we propose {\Ours} for efficient open-vocabulary segmentation in 3D Gaussian scenes.
PairGS shifts the focus from per-Gaussian language or instance features to pairwise relations between Gaussians.
This relation-centric formulation produces cleaner segmentation, including thin structures and contact regions.
{\Ours} further improves efficiency through two-stage relation graph construction and derives a single hierarchical cluster tree that supports multi-granular queries. 
Consequently, {\Ours} achieves outstanding performance while being over $50\times$ faster than optimization-based methods in its fast variant.

\noindent \textbf{Limitations.}
Although PairGS achieves strong performance across diverse benchmarks, it has limitations.
First, like existing approaches~\cite{opengaussian_2024,instancegaussian_2025,cos3d_2025}, it relies on mask evidence for decomposition.
If a small object is never separated in any mask across views, {\Ours} cannot form a cluster for it.
Second, retrieval depends on SAM-masked CLIP~\cite{sam_2023,clip_2021}.
Queries may fail when the cluster features are semantically ambiguous, even with correct geometric clustering.
Further analysis is shown in the supplementary material.

%% file: sec/X_suppl.tex
\clearpage
\setcounter{page}{1}
\appendix

\setcounter{table}{0}
\setcounter{figure}{0}
\renewcommand{\thefigure}{\Alph{figure}}
\renewcommand{\thetable}{\Alph{table}}
\section{Appendix}
{
\makeatletter
\renewcommand*\l@subsection{\@dottedtocline{2}{1.5em}{3.0em}}
\makeatother

\etocsettocstyle{\subsection*{Contents}}{}
\etocsetnexttocdepth{subsection}
\localtableofcontents
}

\subsection{Comparison of Feature Lifting Strategies}
\input{figs/fast_high}
We analyze the trade-off between segmentation quality and computational efficiency by comparing two feature lifting strategies, as shown in~\cref{fig:fast_high}. 
{\Ours}-Fast prioritizes computational efficiency. It performs semantic feature lifting using only the Gaussian centers. In this scenario, the centers of boundary Gaussians are located inside the object. However, the 3D object selection output often extends beyond the actual object boundary. In contrast, {\Ours} considers the contribution of the Gaussian to all pixels. This results in cleaner 3D segmentation.

\subsection{Our 3D--2D Association Approach}
To enable precise open-vocabulary querying, we refine the semantic features of the generated hierarchical clusters. Unlike OpenGaussian~\cite{opengaussian_2024}, our approach bypasses the instance feature similarity filtering, allowing for a efficient implementation.
First, we generate a view-dependent cluster map $\mathcal{M}^{\text{cluster}}_{v}$ using a lightweight ID rasterizer. For each pixel, the rasterizer assigns the ID of the Gaussian with the maximum contribution weight (as defined in~\cref{eq:3dgs_color}), which is subsequently mapped to its cluster label. Background pixels are marked as invalid.

For each cluster $C$, we identify its corresponding dominant 2D SAM mask $\mathbf{m}^{\text{major}}_{v}$ via majority voting, utilizing the per-Gaussian mask vectors. We then validate this association by computing the IoU between the projected cluster region in $\mathcal{M}^{\text{cluster}}_{v}$ and $\mathbf{m}^{\text{major}}_{v}$ to filter out mismatched masks.
Finally, we update the cluster's semantic feature by aggregating the CLIP features of these valid masks. To prioritize high-quality observations, the contribution of view $v$ is weighted by $W_v$:
\begin{equation}
    W_v = \left( \sum_{i \in C \cap \mathbf{m}^{\text{major}}_{v}} \omega_{iv} \right) \times |\mathbf{m}^{\text{major}}_{v}|,
\end{equation}
where the first term represents the total weight of Gaussians corresponding to the mask, and the second term (mask size) reflects the object's scale, ensuring that closer views with richer semantic details contribute more significantly.

Using these view weights, we perform a two-step aggregation. First, we compute the semantic feature $\mathbf{f}^{\text{sem}}_i$ for each Gaussian $i \in C$ by aggregating CLIP features $\mathbf{F}^{\text{sem}}_{v}$ only from views where the Gaussian belongs to the dominant mask:
\begin{equation}
    \mathbf{f}^{\text{sem}}_i = \frac{\sum_{v} \mathbb{I}(m_{iv} = \mathbf{m}^{\text{major}}_{v}) \cdot W_v \cdot \mathbf{F}^{\text{sem}}_{v}}{\sum_{v} \mathbb{I}(m_{iv} = \mathbf{m}^{\text{major}}_{v}) \cdot W_v},
\end{equation}
where $\mathbb{I}(\cdot)$ is an indicator function that ensures consistency with the mask vector $m_{iv}$. Finally, to obtain the unified cluster-level feature $\mathbf{f}^{\text{sem}}_C$, we aggregate these per-Gaussian features based on their opacity $\alpha_i$:
\begin{equation}
    \mathbf{f}^{\text{sem}}_C = \frac{\sum_{i \in C} \alpha_i \mathbf{f}^{\text{sem}}_i}{\sum_{i \in C} \alpha_i}.
\end{equation}
This opacity-weighted averaging ensures that the final cluster embedding is driven by the most visible and structurally significant Gaussians.

\subsection{Our Limitations}
\input{figs/mask_level}
\noindent\textbf{Mask Dependency.}
Prior hierarchical approaches~\cite{langsplat_2024,laga_2025,thgs_2025} use multi-level masks to build a hierarchy.
By using masks at different granularities, they can segment objects from subpart-level clusters up to whole-level clusters.
However, this design often requires optimizing separate semantic or instance features for each mask level, which increases computational cost.

Our method instead leverages a key property of multi-view masks: they are inherently view-inconsistent.
This inconsistency is incorporated into the pairwise affinity in \cref{eq:edge_weight}.
By monotonically increasing the affinity threshold, a hierarchical cluster tree can be constructed from a single mask level.
This strategy improves efficiency, but it introduces a lower bound on the finest granularity that can be represented.
Specifically, extremely fine parts that are separated only at the subpart level, or objects that are never split in any view at the whole-level mask, cannot be identified as meaningful clusters.

For example, the subpart masks in \cref{fig:mask_level} separate the \texttt{pumpkin stem} and some small regions within the \texttt{camera body}.
Under the single-level setting, our method treats these fragments as noise rather than meaningful clusters, and extending granularity while preserving efficiency remains an important direction.

\input{figs/fail_ours}
\noindent\textbf{SAM-masked CLIP Dependency.}
As discussed in the limitations, our method can select an incorrect object for an open-vocabulary query, even when the clusters are geometrically well-formed.
\Cref{fig:fail_ours} shows two representative cases.

In the first example, the ramen contains two eggs, but the prediction for \texttt{egg} retrieves only one.
This error is driven by semantic purity within the hierarchy.
During hierarchical clustering, the parent cluster containing both eggs is split into child clusters.
In some cases, a single-egg child cluster attains higher similarity to the query than the parent cluster that contains both eggs, which leads to an incomplete selection.

In the second example, the query \texttt{paper napkin} matches a visually similar object instead of the ground truth.
This issue is amplified by the evaluation protocol with fixed ground-truth labels, and it can often be alleviated by using a more specific query, such as \texttt{white napkin}.
Finally, in the last row of \cref{fig:fail_ours}, we observe that an incorrect object is selected together with the target for the queries \texttt{stainless steel pots} and \texttt{pour-over vessel}.
While SAM-masked CLIP~\cite{sam_2023,clip_2021,langsplat_2024} features provide dense semantic cues and are widely used in prior works~\cite{drsplat_2025,occams_2024,opengaussian_2024,instancegaussian_2025,laga_2025,cos3d_2025,thgs_2025}, more accurate language embedding and improved inference strategies remain important directions for future research.

\subsection{Evaluating 3D Gaussian Segmentation via Novel View Synthesis}
\input{table/LERF_supp}
Existing evaluation protocols (\cref{tab:lerf}) for open-vocabulary 3D Gaussian segmentation on LERF~\cite{lerf_2023} primarily focus on spatial localization accuracy, typically quantified by mIoU. While mIoU is a robust metric, we argue that evaluating the visual fidelity of segmentation results is equally critical in Gaussian Splatting (GS) domain. Consequently, we incorporate additional evaluations using standard Novel View Synthesis (NVS) quality metrics, where we render each model's output clusters or Gaussian points corresponding to the input language queries. As demonstrated in \cref{tab:lerf_nvs}, our approach achieves superior rendering quality compared to existing baselines. This indicates that our method not only ensures precise locality but also maintains structural cohesiveness, characterized by a dense and consistent point distribution within the segmented regions.

\input{table/scannet200}
\subsection{Results on ScanNet200 dataset}
ScanNet200~\cite{scannet200_2022} dataset extends the label space of ScanNet~\cite{scannet_2017} from 20 to 200 classes, providing a benchmark that better reflects the complexity and semantic diversity of real-world indoor scenes.
It includes many long-tail categories with low occurrence, such as \texttt{dumbbell} or \texttt{lamp}, which makes open-vocabulary understanding particularly challenging.
We evaluate on ScanNet200 using the same evaluation protocol as in \cref{scannet_eval}.
As shown in \cref{tab:scannet200}, our method remains competitive and outperforms prior baselines.

\subsection{Effect of Hierarchical Pruning}
\input{table/Hierarchical_Pruning}
As discussed in \cref{sec:tree_dbscan}, our hierarchical pruning does not include all clusters from every hierarchy level in the tree.
Instead, it retains only meaningful clusters.
\Cref{tab:pruning} shows the number of clusters and the runtime of the subsequent semantic refinement for different hierarchy levels.
Including all clusters causes the number of clusters to grow proportionally with the hierarchy level, while also leading to a slight performance decrease.
In contrast, hierarchical pruning yields only modest changes in the number of retained clusters as the level increases, which reduces the cost of semantic refinement and improves overall efficiency.

\subsection{Runtime for Each Component}
\input{table/time}

\Cref{tab:time} details the runtime analysis for each component of our pipeline. In particular, for {\Ours} with contribution-based initialization, the initialization stage accounts for 71.4\% of the total runtime.
After initialization, the two-stage relation graph construction and hierarchical clustering finish quickly, taking about 20 seconds.
Semantic refinement then takes about 30 seconds.
This is more efficient than the 3D--2D association procedure in OpenGaussian~\cite{opengaussian_2024}, but there remains room for further speedup.

\subsection{Experimental Hyperparameter Details}
\Cref{tab:hyperparams} provides the detailed hyperparameters used in our experiments. Please note that \emph{all the parameters were fixed across all scenes and datasets shown in this paper}.
\input{table/hyperparameter}

\subsection{Hyperparameter Sensitivity}
\input{table/sensitivity}
We further examine the sensitivity of our method to key hyperparameters. As shown in \cref{tab:sensitivity}, we vary the hierarchy level $L$ and digraph out-degree $k$ to $\{20,40\}$, and the PCA dimension $d_{pca}$ to $\{3,24\}$.
The method remains stable as $L$ increases, since hierarchical pruning continues to retain only meaningful clusters.
It also remains stable as $k$ increases, because the digraph design preserves consistent local connectivity.
For PCA dimensionality, the projection primarily serves to provide lightweight semantic cues to node descriptors, so we empirically choose $d_{pca}$.
While performance varies slightly across different settings, it does not collapse under these perturbations and consistently remains superior to other baselines~\cite{langsplat_2024,drsplat_2025,occams_2024, opengaussian_2024,instancegaussian_2025,laga_2025,cos3d_2025,thgs_2025}.

\subsection{Runtime Scalability} 
\cref{tab:scalability} demonstrates the practical scalability of our approach in terms of runtime and memory usage.
We measure the peak VRAM usage during digraph construction and subsequent hierarchical clustering.
\input{table/scalability}

\subsection{Summary of Evaluation Protocol Differences}
\input{figs/2d_eval_supp}
As discussed in the~\cref{sec:experiments}, we follow the 3D evaluation protocol of OpenGaussian~\cite{opengaussian_2024}. 
In the open-vocabulary 3D object selection setting, the evaluation is conducted as follows.
After selecting the 3D clusters corresponding to the query, we render the selected clusters without any additional 2D post-processing and compare the resulting masks with the ground-truth masks.

However, several recent methods~\cite{laga_2025,thgs_2025,relags_2026} additionally apply post-hoc 2D refinement during 3D evaluation. LaGa~\cite{laga_2025} renders 2D similarity maps based on the query similarity scores of Gaussians, and then applies kernel-based smoothing similar to LangSplat~\cite{langsplat_2024}. THGS~\cite{thgs_2025} and its follow-up ReLaGS~\cite{relags_2026} also render 2D score maps from binary Gaussian scores and apply thresholding. Such 2D refinement is effective for removing noisy Gaussians. However, as shown by the cluster result example from ReLaGS~\cite{relags_2026} in~\cref{fig:2d_eval_supp}, it can also create a substantial gap from the original shapes of the model's cluster results, making it less direct to evaluate the intrinsic quality of the underlying 3D clusters.

\input{table/2D_eval_suppl}
\Cref{tab:2d_eval} shows the 3D evaluation results under two protocols: the original OpenGaussian protocol (left, as in ~\cref{tab:lerf}), and the evaluation protocol incorporating each method's additional 2D refinement process (right). PairGS achieves state-of-the-art performance under both evaluation protocols. Moreover, the small numerical gap between the two protocols indicates that the 3D clusters produced by PairGS are clean and less affected by noisy regions even before applying 2D refinement.

We further note that even under the same OpenGaussian~\cite{opengaussian_2024} protocol, we use a different threshold for reporting mAcc. Prior works report mAcc@0.25, while we reported mAcc@0.50 for a more stringent evaluation of segmentation quality, as stated in the caption. For a direct comparison, we report mAcc@0.25 in~\cref{tab:m25} below.
\input{table/m25}

\subsection{Ablation on Clustering Algorithms}
We conduct an ablation by replacing our TreeDBSCAN with HDBSCAN. HDBSCAN builds an MST-based hierarchy from mutual reachability distances and extracts stable clusters, whereas TreeDBSCAN directly thresholds the retained affinities and removes redundant clusters based on parent-child relationships.
\input{table/hdbscan}

As shown in~\cref{tab:hdbscan}, replacing TreeDBSCAN with HDBSCAN leads to a performance drop. This suggests that, on our already-sparsified relation graph, constructing an additional MST-based hierarchy can be less effective than directly exploiting the preserved local affinities. Since our graph construction already filters out weak or noisy relations, TreeDBSCAN can better preserve the intended local connectivity and object-level grouping, leading to cleaner clusters and better downstream performance.

\subsection{Video Tracker Failure Cases}
\input{figs/fail_deva}
Gaussian Grouping~\cite{gaussiangrouping_2024} and ObjectGS~\cite{objectgs_2025} serve as our instance segmentation baselines on the ScanNet~\cite{scannet_2017} dataset. They perform mask association using a video tracker~\cite{deva_2023}. However, this process is highly sensitive to the appearance of new objects and the initial view. As shown in~\cref{fig:fail_deva}, the tracker often fails even with minor viewpoint changes for the same object. This failure leads to the generation of inaccurate pseudo-view-consistent masks. These incorrect pseudo-labels are continuously used during the subsequent optimization process. Consequently, the scene decomposition results are significantly affected by the quality of the mask association.

\subsection{Additional Qualitative Results}
We provide additional qualitative results on the LERF~\cite{lerf_2023}, ScanNet~\cite{scannet_2017} and Mip-NeRF 360~\cite{mipnerf_2022} datasets in~\cref{fig:qual_lerf_supp,fig:qual_scannet_supp,fig:qual_mipnerf_supp}.
We also evaluate open-vocabulary object selection on the outdoor KITTI-360~\cite{kitti360_2022} dataset. \Cref{fig:kitti} shows that our method achieves stable performance even on challenging street scene, outperforming the baselines~\cite{drsplat_2025,opengaussian_2024}.

\subsection{Real-world Application}
As shown in \cref{fig:robot_demo}, we demonstrate a real-world application of {\Ours} for language-driven robot manipulation.
Given the query \texttt{mustard}, we first segment the target 3D Gaussians and convert both the target and background Gaussians into point clouds using their Gaussian centers.
We then use GraspGen~\cite{graspgen_2025} to generate grasp candidates for the target object and execute robot manipulation based on these candidates.
\input{figs/kitti}
\input{figs/hier}

\input{figs/robot_demo}

\input{figs/qual_lerf_supp}

\input{figs/qual_scannet_supp}

\input{figs/qual_mipnerf_supp}

%% file: figs/fast_high.tex
\begin{figure}[h!]
\centering
\includegraphics[width=0.995\columnwidth]{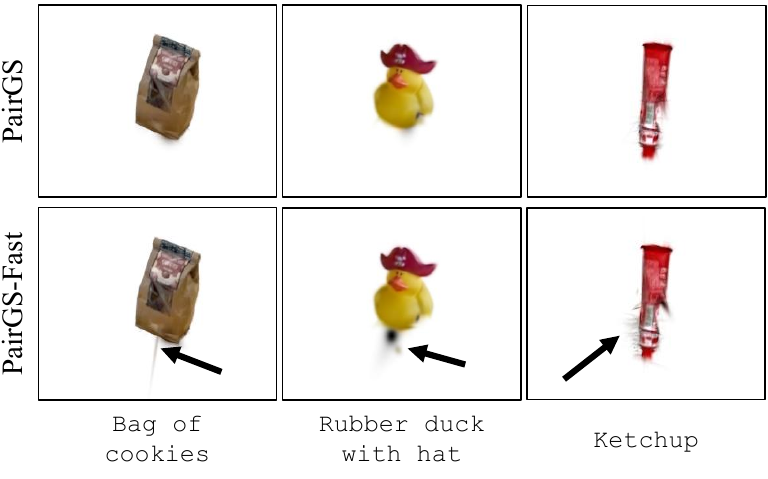}
    \caption{\textbf{Qualitative comparison between {\Ours} and {\Ours}-Fast.} {\Ours}, employing contribution-weighted lifting, produces cleaner and sharper segmentation boundaries. {\Ours}-Fast, utilizing center-based lifting, exhibits boundary bleeding artifacts where the segmentation extends beyond the object.}
    \label{fig:fast_high}
\end{figure}

%% file: figs/mask_level.tex
\begin{figure}[h!]
\centering
\includegraphics[width=1.0\columnwidth]{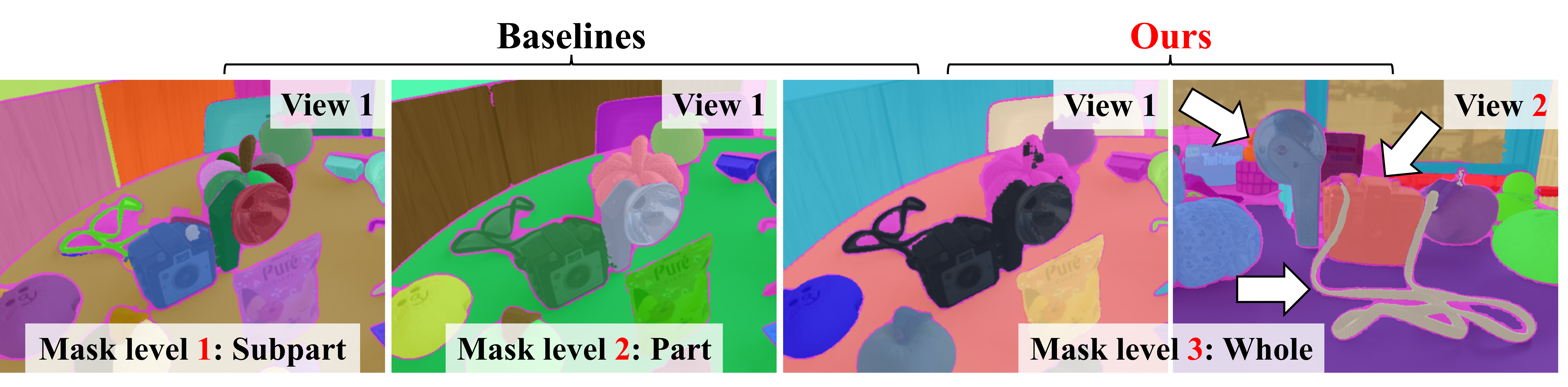}
    \caption{\textbf{Mask Visualization.} SAM~\cite{sam_2023} masks are provided at three granularities: subpart, part, and whole. Hierarchical baselines~\cite{thgs_2025,langsplat_2024,laga_2025} typically use all mask levels to construct a hierarchy. In contrast, we build a hierarchical tree from a single mask level by leveraging view inconsistency across observations. This design cannot form an independent cluster for an object that is never separated in any other view.}
    \label{fig:mask_level}
\end{figure}

%% file: figs/fail_ours.tex
\begin{figure}[h!]
\centering
\includegraphics[width=0.995\columnwidth]{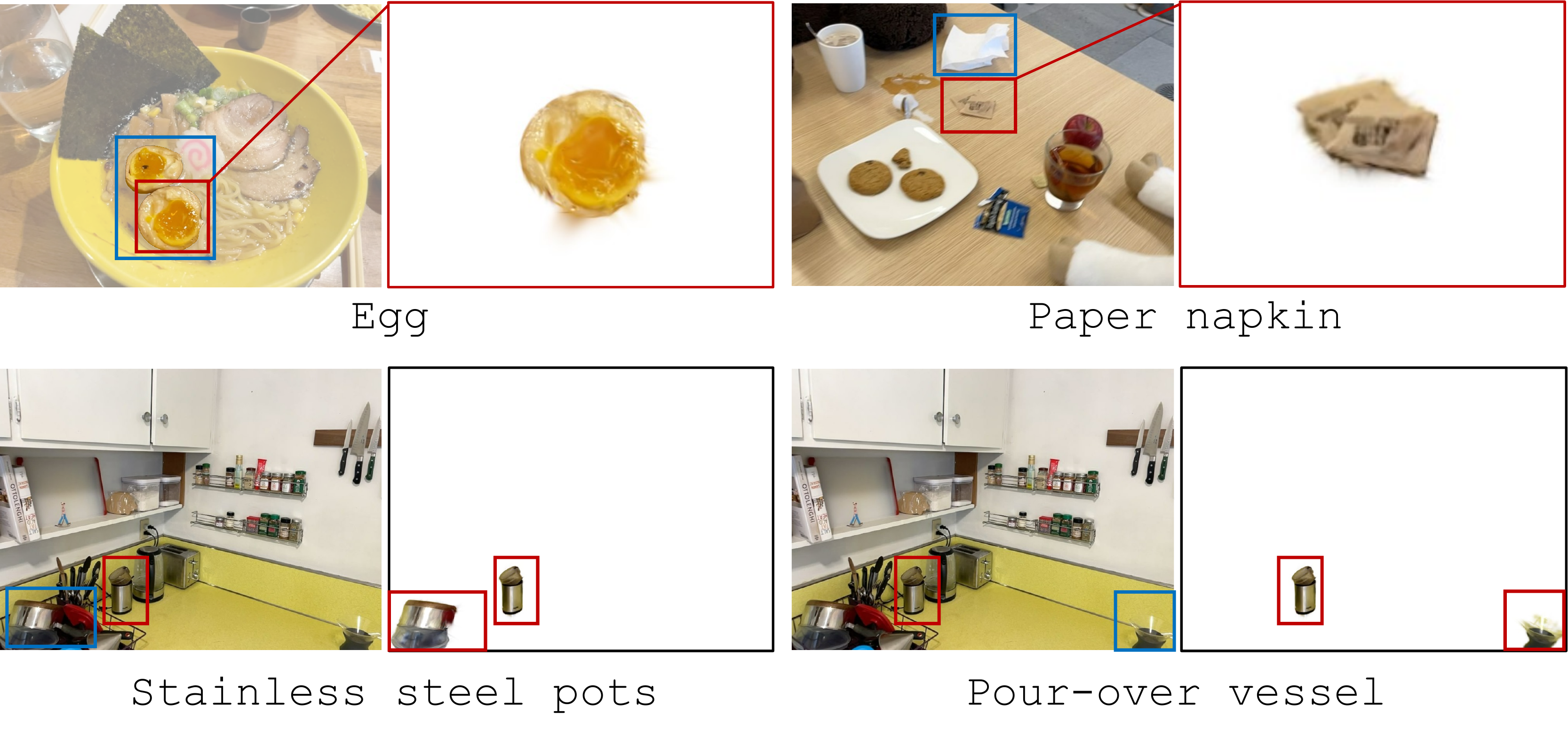}
    \caption{\textbf{Failure cases.} The blue boxes indicate the ground truth masks. The red boxes represent the predictions.}
    \label{fig:fail_ours}
\end{figure}

%% file: table/LERF_supp.tex
\begin{table}[h!]
\vspace{-3mm}
\caption{Rendering quality evaluation LERF~\cite{lerf_2023} dataset}
\centering
\resizebox{0.99\columnwidth}{!}{
\setlength{\tabcolsep}{8pt}
\begin{tabular}{l|ccc}
\toprule[1.2pt]
& \textbf{PSNR $\uparrow$} & \textbf{SSIM $\uparrow$} & \textbf{LPIPS $\downarrow$} \\
\midrule
LangSplat~\cite{langsplat_2024} & 17.87 & 0.864 & 0.270 \\
Occam's LGS~\cite{occams_2024} & 22.23 & 0.934 & 0.134 \\
Dr.Splat~\cite{drsplat_2025} & \cellcolor{colort}24.90 & 0.955 & 0.093 \\
\midrule
OpenGaussian~\cite{opengaussian_2024} & 24.82 & 0.966 & \cellcolor{colort}0.062 \\
InstanceGaussian~\cite{instancegaussian_2025} & 22.84 & 0.871 & 0.080 \\
COS3D~\cite{cos3d_2025} & 24.22 & 0.958 & 0.084 \\
LaGa~\cite{laga_2025} & 24.64 & 0.959 & 0.074 \\
\midrule
THGS~\cite{thgs_2025} & 23.70 & \cellcolor{colort}0.969 &  \cellcolor{colort}0.062 \\
{\Ours}-Fast (Ours) & \cellcolor{colors}27.60 & \cellcolor{colors}0.972 & \cellcolor{colors}0.056 \\
{\Ours} (Ours) & \cellcolor{colorf}28.10 & \cellcolor{colorf}0.976 & \cellcolor{colorf}0.046 \\
\bottomrule[1.2pt]
\end{tabular}
}
\label{tab:lerf_nvs}
\end{table}

%% file: table/scannet200.tex
\begin{table*}[t!]
\vspace{-3mm}
\centering
\small
\caption{Open-vocabulary scene understanding results on the ScanNet200~\cite{scannet200_2022} dataset}
\resizebox{0.9\textwidth}{!}{
\setlength{\tabcolsep}{6pt}
\begin{tabular}{
    l | c c c c c c| c c
}
\toprule[1.2pt]
&Occam's~\cite{occams_2024}  & Dr.Splat~\cite{drsplat_2025} & OpenGS~\cite{opengaussian_2024} & InstanceGS~\cite{instancegaussian_2025} &LaGa~\cite{laga_2025} & THGS~\cite{thgs_2025} & {\Ours}-Fast & {\Ours}\\
\midrule
 mIoU$\uparrow$ & 16.29 &13.37&13.43&19.76 &14.95&18.45&\underline{20.80}&\textbf{22.06}  \\
 mAcc$\uparrow$ &22.15&18.38&18.91&27.87 &25.01&27.60&\underline{28.73}&\textbf{29.89} \\
\bottomrule[1.2pt]
\end{tabular}
}
\label{tab:scannet200}
\end{table*}

%% file: table/Hierarchical_Pruning.tex
\begin{table*}[t!]
\centering
\caption{Effect of hierarchy level on cluster count and semantic refinement runtime}
\resizebox{0.95\textwidth}{!}{
\setlength{\tabcolsep}{6pt}
\renewcommand{\arraystretch}{1.2}
\begin{tabular}{
    l | c c c | c c c | c c c
}
\toprule[1.2pt]
& 
\multicolumn{3}{c|}{\textbf{Level=10}} &
\multicolumn{3}{c|}{\textbf{Level=20}} &
\multicolumn{3}{c}{\textbf{Level=30}}\\
& \# Clusters &SR time$\downarrow$ (s) &mIoU
& \# Clusters &SR time$\downarrow$ (s) &mIoU
& \# Clusters &SR time$\downarrow$ (s) &mIoU\\
\midrule
Ours(TreeDBSCAN)
&{\large \textbf{155.8}} &{\large \textbf{30.3}} &{\large\textbf{60.4}}
&{\large \textbf{172.8}} &{\large \textbf{44.8}} &{\large\textbf{59.8}}
&{\large \textbf{188.5}} &{\large \textbf{58.5}} &\textbf{\large{60.3}}\\
w/o Hierarchical Pruning 
&{\large 650.5{\small(\textbf{\textcolor{softred}{$\times$4.18}})}} &{\large 59.41{\small(\textbf{\textcolor{softred}{$\times$1.96}})}} &{\large{57.3}}
&{\large 1326.3{\small(\textbf{\textcolor{softred}{$\times$7.68}})}} &{\large 115.1{\small(\textbf{\textcolor{softred}{$\times$2.57}})}} &\large{{58.3}}
&{\large 2001.5{\small(\textbf{\textcolor{softred}{$\times$10.62}})}} &{\large 174.6{\small(\textbf{\textcolor{softred}{$\times$2.98}})}} &{\large{59.5}}\\
\bottomrule[1.2pt]
\end{tabular}
}
\label{tab:pruning}
\end{table*}

%% file: table/time.tex
\begin{table}[h!]
\vspace{-4mm}
\caption{\textbf{Runtime per Components on LERF~\cite{lerf_2023} dataset.} The table shows the average runtime (in seconds) required for Initialization (IN), Sparse Edge Proposal (SP), Pairwise Affinity Formulation (PA), Hierarchical Clustering(HC), and Semantic Refinement (SR).}
\centering
\resizebox{0.9\columnwidth}{!}{
\setlength{\tabcolsep}{5pt}
\begin{tabular}{
    l | c c c c c c
}
\toprule[1.2pt]
Method & IN & SP & PA & HC & SR & \textbf{Total} \\
\midrule
{\Ours}-Fast & \multicolumn{1}{r}{10.30} & \multicolumn{1}{r}{14.87} & \multicolumn{1}{r}{0.34} & \multicolumn{1}{r}{4.55} & \multicolumn{1}{r}{28.10} & \multicolumn{1}{r}{58.16} \\
{\Ours} & \multicolumn{1}{r}{130.13} & \multicolumn{1}{r}{17.42} & \multicolumn{1}{r}{0.45} & \multicolumn{1}{r}{3.97} & \multicolumn{1}{r}{30.27} & \multicolumn{1}{r}{182.24} \\
\bottomrule[1.2pt]
\end{tabular}
}
\label{tab:time}
\end{table}

%% file: table/hyperparameter.tex
\begin{table}[h!]
    \centering
    \caption{List of hyperparameters used in our method}
    \label{tab:hyperparams}
    \resizebox{0.9\columnwidth}{!}{
    \renewcommand{\arraystretch}{1.2}
    \begin{tabular}{lcc}
        \toprule
        \textbf{Parameter} & \textbf{Symbol} & \textbf{Value} \\
        \midrule
        \multicolumn{3}{l}{\textit{\textbf{Sparse Edge Proposal}}} \\
        PCA dimension & $d_{pca}$ & 6 \\  
        Graph neighbors ($k$-NN) & $k$ & 10 \\
        \midrule
        \multicolumn{3}{l}{\textit{\textbf{Hierarchical Clustering}}} \\
        Number of hierarchy levels & $L$ & 10 \\
        Affinity threshold range & $[\theta_{\min}, \theta_{\max}]$ & $[0.6, 0.99]$ \\
        Erosion rate threshold & $\rho_{\min}$ & 0.5 \\
        \bottomrule
    \end{tabular}
    }
\end{table}

%% file: table/sensitivity.tex
\begin{table}[h!]
    \centering
    \vspace{-3mm}
    \caption{Hyperparameter sensitivity results.}
    \label{tab:sensitivity}
    \resizebox{0.9\linewidth}{!}{
    \setlength{\tabcolsep}{8pt}
    \renewcommand{\arraystretch}{1.2}
    \begin{tabular}{l | c c | c c}
    \toprule[1.2pt]
    &
    \multicolumn{2}{c|}{\textbf{LERF}~\cite{lerf_2023}} &
    \multicolumn{2}{c}{\textbf{ScanNet}~\cite{scannet_2017}}\\
    Case & mIoU$\uparrow$ & mAcc$\uparrow$ & mIoU$\uparrow$ & mAcc$\uparrow$\\
    \midrule
    $L = 20$        & 59.8 & 67.8 & 51.4 & 66.6\\
    $L = 40$        & 59.6 & 67.4 & 50.1 & 65.6\\
    \midrule
    $k = 20$        & 59.7 & 65.8 & 54.9 & 72.6\\
    $k = 40$        & 59.0 & 65.4 & 55.5 & 71.3\\
    \midrule
    $d_{pca} = 3$        & 60.2  & 67.4 & 52.2 & 66.6\\
    $d_{pca} = 24$        & 57.1 & 62.8 & 55.1 & 70.2\\
    \midrule
    
    default & 60.4 & 68.2 & 53.0 & 68.1\\
    \bottomrule[1.2pt]
    \end{tabular}
    }
\end{table}

%% file: table/scalability.tex
\begin{table}[h!]
\centering
\caption{Scalability results.}
\label{tab:scalability}
\footnotesize
\setlength{\tabcolsep}{4pt}
\renewcommand{\arraystretch}{1.05}
\begin{tabular}{l|cc|c}
\toprule[1.2pt] 
\multirow{2}{*}{\# Gaussians} 
& \multicolumn{2}{c|}{Runtime (s)} 
& \multicolumn{1}{c}{VRAM (MB)} \\
& {\Ours}-Fast & {\Ours} & {\Ours} \\
\midrule
Small ($<$ 0.3M)        & 0:15 & 0:27 & \num{2826} \\
Medium (0.3M--1.0M)    & 0:35 & 1:57 & \num{11179} \\
Large (2.0M--2.5M)     & 1:10 & 3:48 & \num{14723} \\
\bottomrule[1.2pt]
\end{tabular}
\end{table}

%% file: figs/2d_eval_supp.tex
\begin{figure}[h!]
\centering
\vspace{-3mm}
\includegraphics[width=1.0\linewidth]{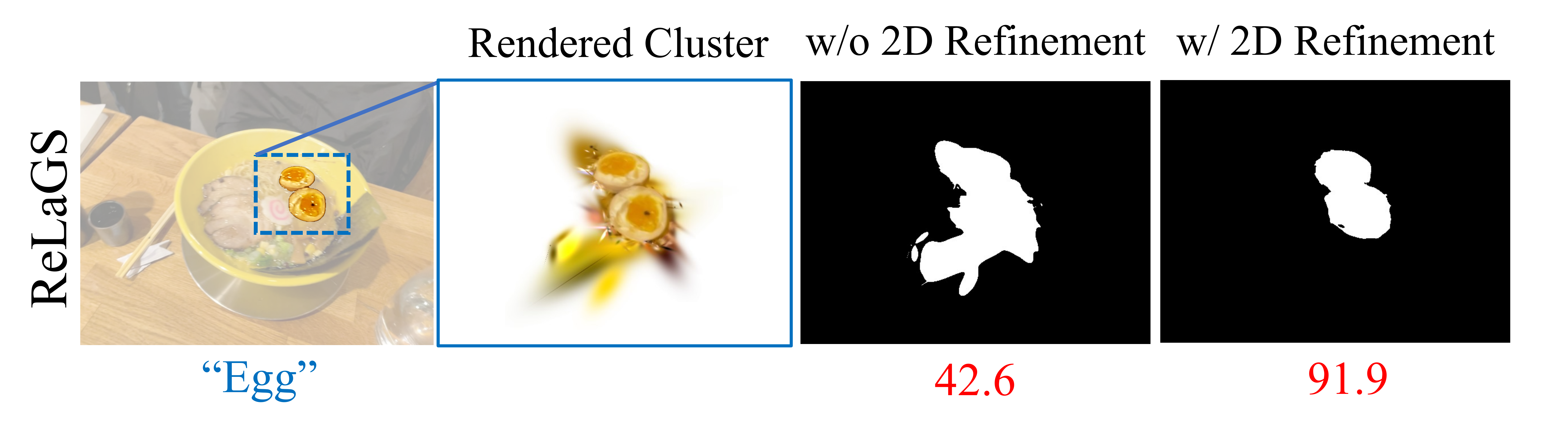}
    \caption{\textbf{Effect of post-hoc 2D refinement in 3D evaluation.} Although 2D refinement suppresses noisy regions during evaluation, it can lead to a substantial mIoU gap between the model’s raw cluster masks (42.6) and the refined masks (91.9) used for comparison.}
    \label{fig:2d_eval_supp}
\end{figure}

%% file: table/2D_eval_suppl.tex
\begin{table}[h!]
\centering
    \caption{Effect of 2D refinement on 3D segmentation performance.}
    \label{tab:2d_eval}
    \small
    \setlength{\tabcolsep}{6pt}
    \resizebox{0.995\columnwidth}{!}{%
    \begin{tabular}{l| c c | c c}
    \toprule[0.8pt]
    &
    \multicolumn{2}{c|}{\textbf{3D}} &
    \multicolumn{2}{c}{\textbf{3D + 2D Refinement}} \\
    Method & mIoU$\uparrow$ & mAcc$\uparrow$ & mIoU$\uparrow$ & mAcc$\uparrow$ \\
    \midrule
    LaGa~\cite{laga_2025} & 52.2 & 55.1 & 60.1 {\scriptsize (64.0$^\dagger$)} & 64.5 \\
    THGS~\cite{thgs_2025} & 42.5 & 43.3 & 56.9 {\scriptsize (54.9$^\dagger$)} & 61.1 \\
    ReLaGS~\cite{relags_2026} & 42.7 & 48.1 & 59.2 {\scriptsize (64.4$^\dagger$)} & 65.4 \\
    PairGS & \textbf{60.4} & \textbf{68.2} & \textbf{64.7} & \textbf{71.7} \\
    \bottomrule[0.8pt]
    \end{tabular}%
    }
    
    \vspace{1mm}
    \footnotesize{$^\dagger$ denotes the result reported in the original paper.}
\end{table}

%% file: table/m25.tex
\begin{table}[h!]
\centering
\caption{Evaluation results on LERF~\cite{lerf_2023} using mAcc@0.25.}
\resizebox{0.995\columnwidth}{!}{
\setlength{\tabcolsep}{5pt}
\begin{tabular}{
    l | c c c c c c c| c
}
\toprule[1.2pt]
&\shortstack{Occam's\\[-0.5mm]{\cite{occams_2024}}}  & \shortstack{Dr.Splat\\[-0.5mm]{\cite{drsplat_2025}}} & \shortstack{OpenGS \\[-0.5mm]{\cite{opengaussian_2024}}}& \shortstack{InstanceGS \\[-0.5mm]{\cite{instancegaussian_2025}}}&\shortstack{COS3D \\[-0.5mm]{\cite{cos3d_2025}}}&\shortstack{LaGa \\[-0.5mm]{\cite{laga_2025}}}& \shortstack{THGS \\[-0.5mm]{\cite{thgs_2025}}}& \shortstack{PairGS \\[0.5mm]}\\
\midrule
 mAcc@0.25$\uparrow$ &  \large 70.0 &\large 69.5 &\large 58.5 &\large 60.3 &\large 69.6 &\large 76.4 &\large 71.4 &\large \textbf{79.6} \\
\bottomrule[1.2pt]
\end{tabular}
}
\label{tab:m25}
\end{table}

%% file: table/hdbscan.tex
\begin{table}[h!]
\centering
\small
\caption{Ablation study on clustering algorithms.}
\label{tab:hdbscan}
\setlength{\tabcolsep}{5pt}
\resizebox{0.7\columnwidth}{!}{
\begin{tabular}{l | c c}
\toprule[1.2pt]
& TreeDBSCAN & HDBSCAN \\
\midrule
LERF (mIoU$\uparrow$)    &  \textbf{60.4} &  51.7 \\
ScanNet (mIoU$\uparrow$) &  \textbf{40.7} &  38.9 \\
\bottomrule[1.2pt]
\end{tabular}
}
\end{table}

%% file: figs/fail_deva.tex
\begin{figure}[h!]
\centering
\vspace{-3mm}
\includegraphics[width=0.995\columnwidth]{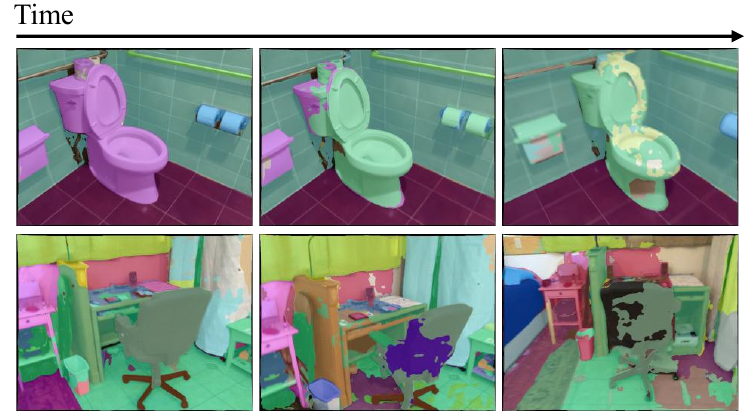}
    \caption{\textbf{Failure case of view-association on ScanNet~\cite{scannet_2017} dataset.} This visualization shows the associated mask IDs overlaid on the image. The same mask color indicates the same object ID. As shown, the video tracker~\cite{deva_2023} fails to maintain consistent identities across different views.}
    \label{fig:fail_deva}
\end{figure}

%% file: figs/kitti.tex
\begin{figure*}[h!]
\centering
\includegraphics[width=0.8\textwidth]{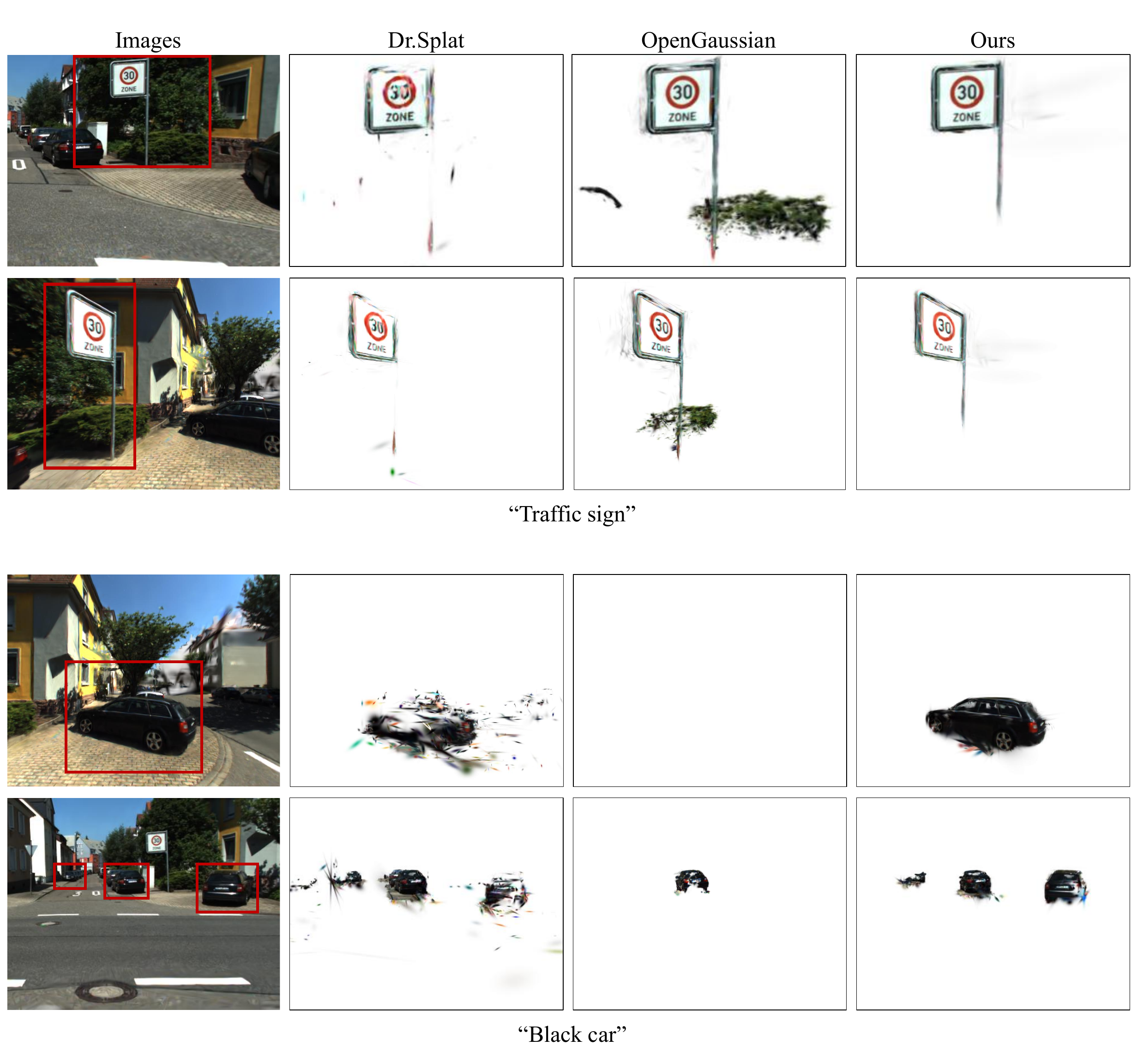}
    \caption{\textbf{Additional Result on KITTI 360~\cite{kitti360_2022} Dataset.}}
    \label{fig:kitti}
\end{figure*}

%% file: figs/hier.tex
\begin{figure*}[h!]
\centering
\vspace{-4mm}
\includegraphics[width=0.95\textwidth]{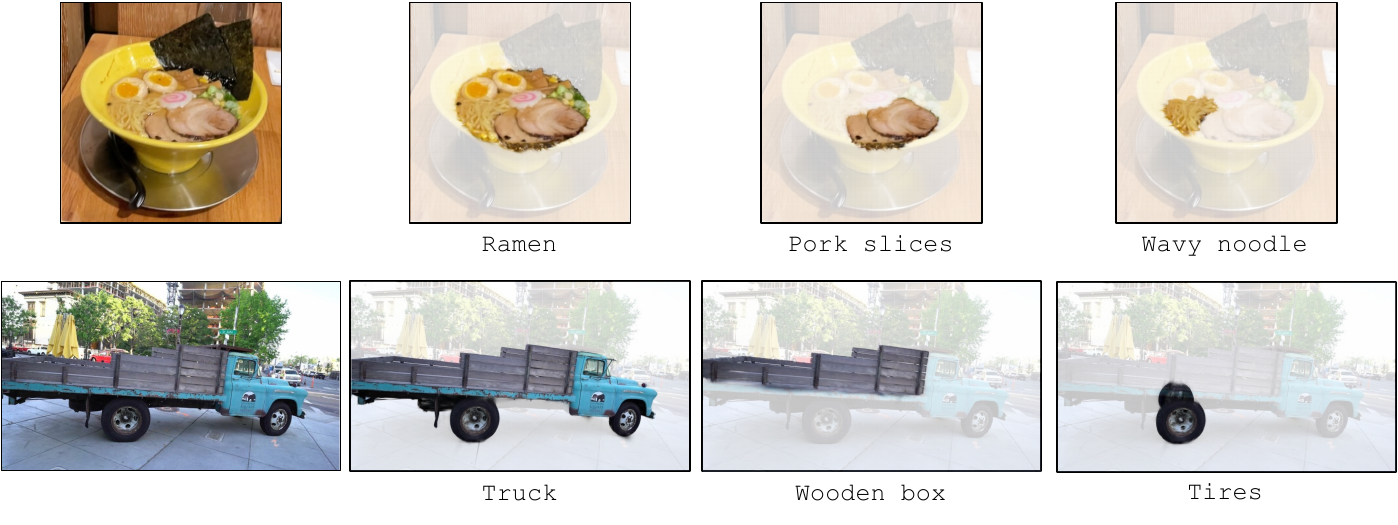}
    \caption{\textbf{Additional multi-granular examples on LERF~\cite{lerf_2023} and Tanks\&Temples~\cite{tank_2017} Dataset.}}
    \label{fig:hier}
\vspace{-4mm}
\end{figure*}

%% file: figs/robot_demo.tex
\begin{figure*}[h!]
\centering
\vspace{-4mm}
\includegraphics[width=0.99\textwidth]{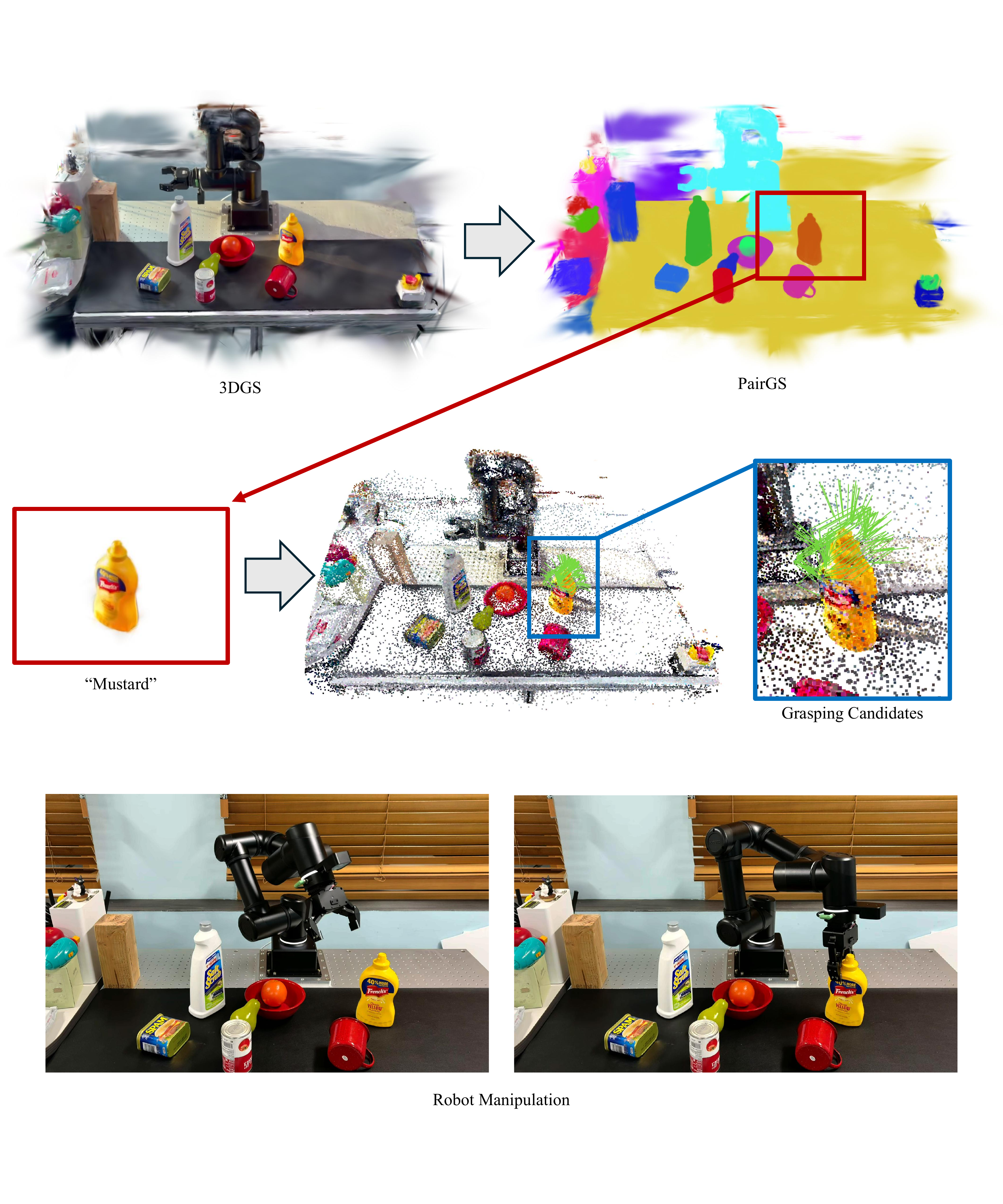}
    \caption{\textbf{Real-world application.}}
    \label{fig:robot_demo}
\end{figure*}

%% file: figs/qual_lerf_supp.tex
\begin{figure*}[h!]
\centering
\vspace{4mm}
\includegraphics[width=0.92\textwidth]{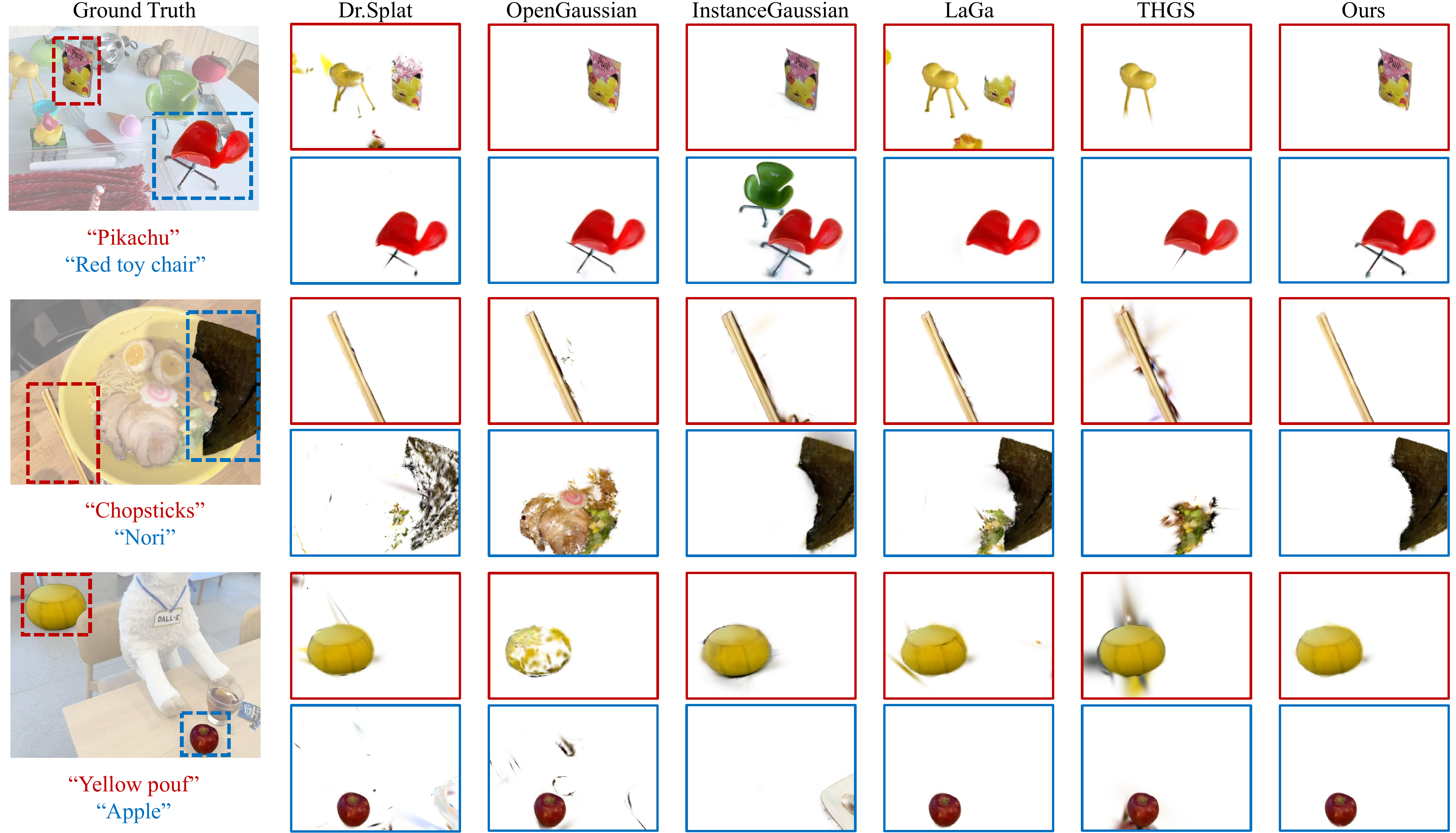}
    \caption{\textbf{Additional Result on LERF~\cite{lerf_2023} Dataset.}}
    \label{fig:qual_lerf_supp}
\end{figure*}

%% file: figs/qual_scannet_supp.tex
\begin{figure*}[t!]
\centering
\vspace{4mm}
\includegraphics[width=0.92\textwidth]{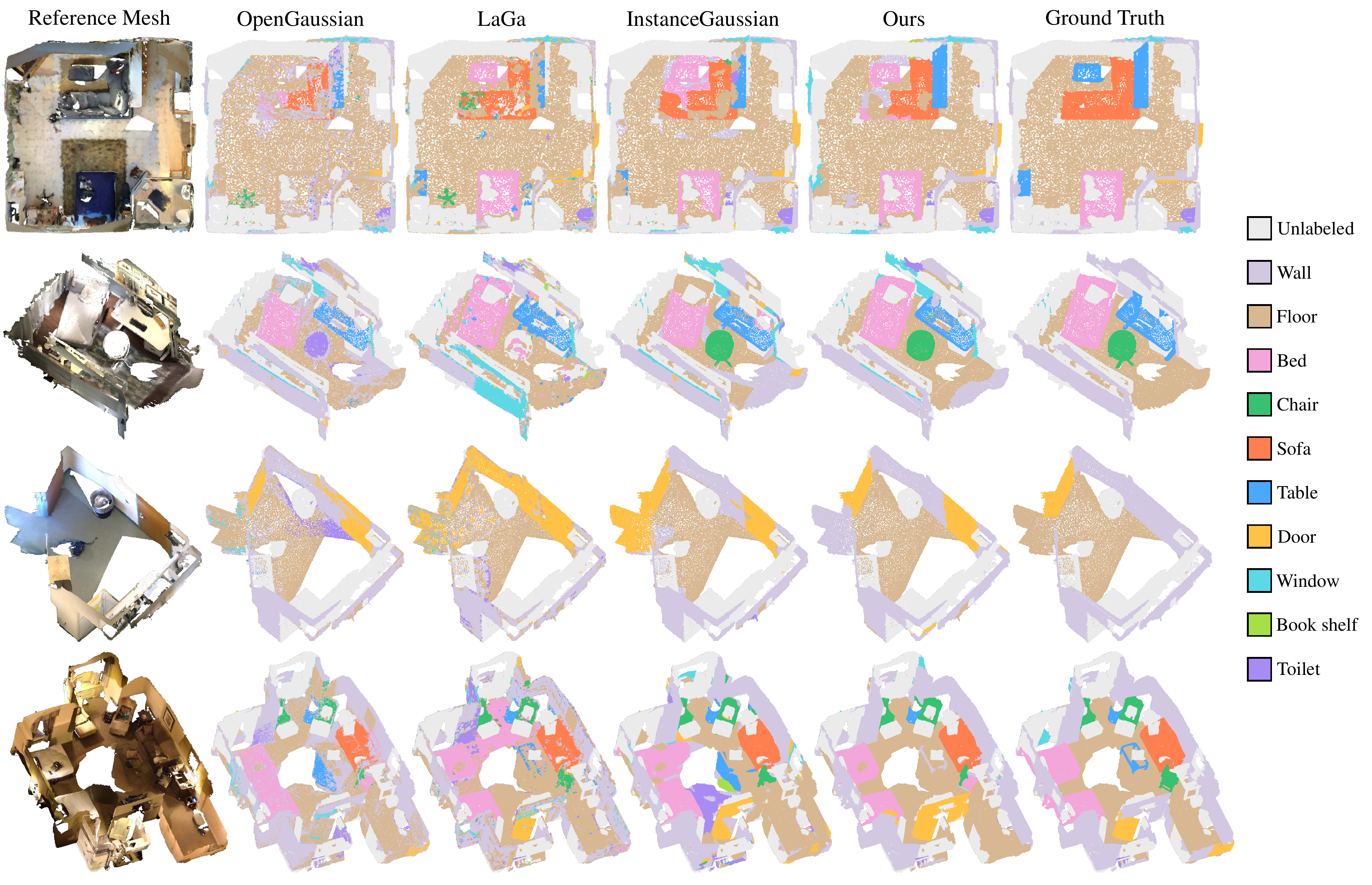}
    \caption{\textbf{Additional Result on Scannet~\cite{scannet_2017} Dataset.}}
    \label{fig:qual_scannet_supp}
\end{figure*}

%% file: figs/qual_mipnerf_supp.tex
\begin{figure*}[h!]
\centering
\vspace{-6mm}
\includegraphics[width=0.999\textwidth]{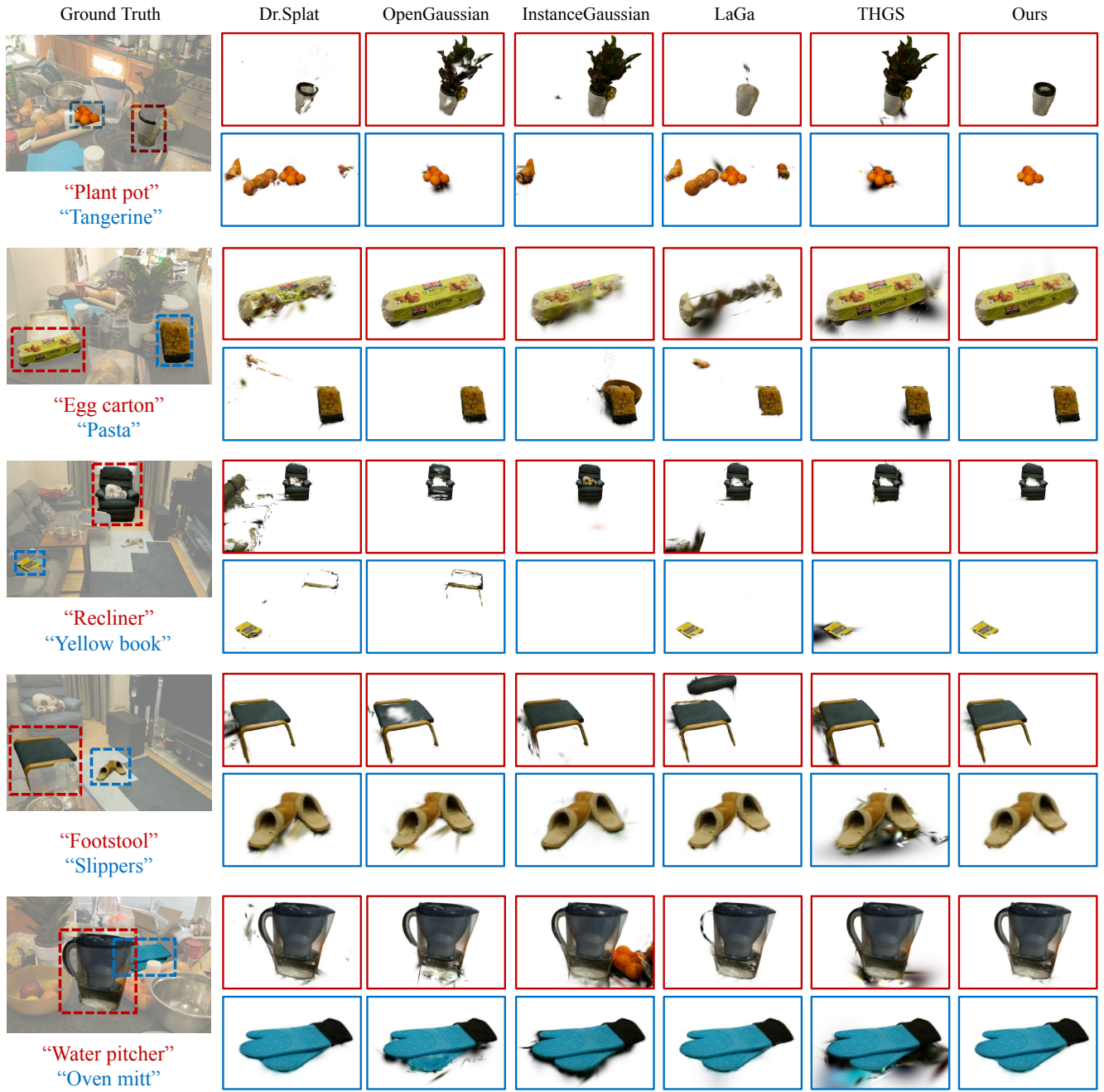}
    \caption{\textbf{Additional Result on Mip-NeRF 360~\cite{mipnerf_2022} Dataset.}}
    \label{fig:qual_mipnerf_supp}
\end{figure*}